%% file: main.tex
\documentclass[10pt,twocolumn,letterpaper]{article}
\usepackage[T1]{fontenc}

\usepackage[pagenumbers]{template/cvpr}
\input{template/preamble}
\usepackage{tikz}
\usetikzlibrary{arrows.meta,backgrounds,calc,decorations.pathreplacing,fit,patterns,positioning}

\definecolor{cvprblue}{rgb}{0.21,0.49,0.74}
\usepackage[pagebackref,breaklinks,colorlinks,allcolors=cvprblue]{hyperref}

\def\confName{CVPR}
\def\confYear{2027}

\title{Proposal2Adapter: Aggregating Instance-Conditioned Low-Rank Updates for Few-Shot Test-Time Domain Adaptation}
\author{Adhitya Kamakshidasan$^{1,2}$ \quad Xiaoya Tang$^{1,2}$ \quad Tolga Tasdizen$^{1,3}$\\
$^{1}$Scientific Computing and Imaging Institute\\
$^{2}$Kahlert School of Computing \quad $^{3}$Department of Electrical and Computer Engineering\\
University of Utah, Salt Lake City, USA}

\begin{document}
\maketitle

\input{sections/abstract}

\input{sections/introduction}
\input{sections/related_work}
\input{sections/method}

\input{sections/analysis}

\input{sections/experiments}

\input{sections/limitations}

\input{sections/conclusion}

\input{sections/acknowledgments}
\input{sections/llm_usage}

{
  \small
  \bibliographystyle{template/ieeenat_fullname}
  \bibliography{references}
}

\end{document}

\maketitle

\setlength{\textfloatsep}{8pt plus 2pt minus 2pt}
\setlength{\dbltextfloatsep}{8pt plus 2pt minus 2pt}
\setlength{\floatsep}{6pt plus 2pt minus 2pt}
\setlength{\dblfloatsep}{6pt plus 2pt minus 2pt}

\input{sections/supplementary_overview}
\input{sections/supplementary_analysis}
\input{sections/supplementary_experiments}
\raggedbottom
\input{sections/supplementary_libero}

{
  \small
  \bibliographystyle{template/ieeenat_fullname}
  \bibliography{references}
}

%% file: template/preamble.tex
\renewcommand{\paragraph}[1]{\vspace{.5em}\noindent\textbf{#1}}

%% file: sections/abstract.tex
\begin{abstract}
Few-shot test-time domain adaptation seeks to adapt a pretrained model to a new domain using only a small unlabeled support set. Existing approaches often require test-time optimization or construct target updates by combining a pre-fitted collection of source-domain adapters. We introduce Proposal2Adapter, an amortized, gradient-free framework that generates a target-domain adapter through a hypernetwork.
We study two permutation-invariant synthesis rules. \emph{Decode--aggregate} decodes one low-rank update proposal per support example and averages the completed updates, whereas \emph{aggregate--decode} first averages support codes and then decodes a single update. With linear factor decoders, decode--aggregate depends on the empirical second moment of the support codes and can represent updates of rank up to $|\mathcal S|r$, while aggregate--decode depends on the mean code and remains rank at most $r$. 
We train episodically across labeled source domains and evaluate on iWildCam, FMoW, and PovertyMap from WILDS, as well as held-out manipulation tasks from LIBERO. On iWildCam, aggregate--decode achieves $39.22$ macro-F1 with ViT-B/16, the strongest result among the compared same-backbone methods. Controlled studies show that OOD performance and support-dependent specialization improve as source supervision becomes richer and better aligned with support-identifiable variation. 
These results establish direct support-to-adapter generation as a scalable approach to few-shot test-time adaptation and reveal how aggregation order and source-domain richness shape adapter capacity and support-dependent specialization.
\end{abstract}

%% file: sections/introduction.tex
\section{Introduction}
\label{sec:introduction}

When machine learning systems are deployed, they often encounter new
distributions that significantly differ from those seen during training,
thereby degrading otherwise strong predictive
performance~\cite{koh2021wilds}. Test-time adaptation (TTA) addresses this
issue by adapting models using unlabeled data from target domains encountered
at inference time~\cite{wang2021tent}. Existing TTA methods assume access
to large unlabeled target datasets~\cite{liang2020shot}
or require repeated gradient-based updates over an incoming
stream~\cite{wang2022cotta}. However, such
requirements are difficult to satisfy when adaptation needs to be immediate,
stable, and computationally inexpensive.

Few-shot test-time domain adaptation (FSTT-DA)~\cite{reid2026damerge} considers a more constrained
setting: a model is first trained using labeled data from multiple source domains,
and then adapts to unseen target domains by using small unlabeled support sets. Such adaptation is useful across many applications. For example, a medical model may need to adapt to data from a new
hospital with different acquisition conditions using a small set of unlabeled
scans. Similarly, robot embodiments may need to adapt to new task requirements using
only a few demonstrations.

In this work, we focus on amortized conditional adaptation~\cite{zhang2021arm},
where episodic source training learns a mapping from an unlabeled target
support set to an adapter that can be applied without inference-time
optimization. Rather than combining adapters learned for source
domains~\cite{reid2026damerge}, a source-shared hypernetwork synthesizes
low-rank factors directly from target-support representations. We study two
orders for combining the support set. \emph{Decode--aggregate} decodes a
low-rank update proposal from each support example before averaging the
completed updates. \emph{Aggregate--decode} instead averages the support codes
before decoding a single update. Both produce one permutation-invariant
target-domain adapter shared across queries.

We introduce \emph{Proposal2Adapter}, with decode--aggregate as its canonical
form. Each support example contributes a compact low-rank proposal, and the
completed proposals are combined into one target-domain adapter. Shared
proposal prediction keeps the trainable parameter count fixed as source
domains are added. We analyze the two aggregation orders in
Section~\ref{sec:aggregation-analysis}. For decode--aggregate, an exact
decomposition isolates the contribution of support variation omitted by
aggregate--decode, and combining $S$ rank-$r$ proposals can yield an adapter
of rank up to $Sr$ when code rank and decoder dimensions permit.
Aggregate--decode depends only on the mean support code and remains rank at
most $r$. These complementary inductive biases motivate evaluating both rules
under standard and increasingly rich source-domain supervision.

\begin{samepage}
We evaluate both aggregation orders on WILDS~\cite{koh2021wilds} and use the
canonical decode--aggregate configuration for the source-domain scaling and
LIBERO studies. On iWildCam, aggregate--decode reaches
$39.22$ macro-F1 with ViT-B/16, exceeding the strongest compared
same-backbone baseline, L2C~\cite{chi2025l2c}, by $4.02$ points.
Decode--aggregate reaches $37.82$. 
In a controlled oracle-label experiment, adding 1,264 source cameras improves both aggregation rules relative to training on the official labeled source set alone.
Decode--aggregate serves as the canonical configuration for our detailed
source-camera and domain-granularity studies, which show that performance and
the benefit of matched target support increase as source variation becomes
richer and aligned with support-identifiable factors. Both aggregation orders
retain approximately $97\%$ of their 16-shot macro-F1 with four support
images. We further extend
image-conditioned adaptation to the LIBERO robot-manipulation
benchmark~\cite{liu2023libero}; with one visual demonstration,
Proposal2Adapter raises LIBERO-Spatial success to $17.87\%$, compared with
$4.20\%$ for source-trained conventional LoRA.

\end{samepage}

\paragraph{Contributions.}
Our contributions are threefold:
\begin{itemize}
  \item We introduce Proposal2Adapter, a source-shared, gradient-free
  hypernetwork that generates one target-domain adapter from unlabeled support
  examples. We study decode--aggregate and aggregate--decode as two
  permutation-invariant synthesis rules with different capacity and
  regularization properties.
  \item We characterize the information and rank capacity associated with
  aggregation order. Decode--aggregate depends on the empirical second moment
  of learned support codes, retains a support-variation term, and can exceed
  the rank of an individual proposal; aggregate--decode depends on the mean
  code and remains rank at most $r$. Measurements of the realized target-domain
  updates confirm this predicted rank separation.
  \item We evaluate Proposal2Adapter on WILDS and extend image-conditioned
  adaptation to robot control on LIBERO. We obtain strong standard iWildCam
  performance and use controlled experiments to study target
  support, conditional specialization, and the scale and structure of source
  supervision.
\end{itemize}

%% file: sections/related_work.tex
\section{Related Work}
\label{sec:related-work}

\paragraph{Test-time adaptation.}
Most TTA methods update model parameters using unlabeled target data. TENT~\cite{wang2021tent} updates normalization parameters by
minimizing prediction entropy, while MEMO~\cite{zhang2022memo} minimizes the
entropy of predictions averaged across augmented views of a test example.
CoTTA~\cite{wang2022cotta} addresses adaptation over a changing target stream,
and EATA~\cite{niu2022eata} reduces unnecessary updates while limiting
forgetting. These methods optimize model parameters using target data.
ARM~\cite{zhang2021arm} takes a different approach, using episodic source
training to learn adaptation from unlabeled target examples. Proposal2Adapter
likewise learns to generate adapters from small support sets without
optimization at inference time.

\paragraph{Few-shot test-time domain adaptation.}
Few-shot test-time domain adaptation (FSTT-DA) limits adaptation to a small
unlabeled support set before processing target queries. Existing methods
transfer knowledge across source domains using this support.
MABN~\cite{wu2024mabn} adapts normalization parameters through a
self-supervised auxiliary branch, while Meta-DMoE~\cite{zhong2022metadmoe}
distills knowledge from multiple source-domain experts into a student model.
VDPG~\cite{chi2024vdpg} generates domain-specific visual prompts for
frozen CLIP, and L2C~\cite{chi2025l2c} complements frozen CLIP with learned
features and domain prompts. Closely related to our work,
DA-MergeLoRA~\cite{reid2026damerge} trains a
LoRA adapter for each source domain and predicts merging weights from target
support. Instead of merging stored source adapters, Proposal2Adapter
combines proposals predicted from individual target support examples into
one adapter.

\paragraph{Conditional parameter generation and adapter composition.}
Hypernetworks~\cite{ha2017hypernetworks} use one network to generate the
parameters of another. CNAPs~\cite{requeima2019cnaps} applies this idea to
few-shot classification, using labeled support examples to predict parameters
that adjust feature extraction. LoRA~\cite{hu2022lora} keeps adaptation compact
by representing weight updates in a low-rank form. Conditional generation also
supports personalization and task adaptation.
HyperDreamBooth~\cite{ruiz2024hyperdreambooth} predicts personalization weights
from an image before refining them through finetuning, while
Text-to-LoRA~\cite{charakorn2025texttolora} generates adapters from natural
language task descriptions. Adapter composition methods, such as
LoRAHub~\cite{huang2024lorahub}, combine adapters trained on existing tasks.
Proposal2Adapter generates low-rank proposals from individual unlabeled support
examples and combines the completed weight updates. The adapter can therefore
express more adaptation directions than a single low-rank proposal.

\paragraph{Robot policy adaptation.}
Pretrained robot policies adapt through finetuning, demonstration
conditioning, and parameter generation.
Octo~\cite{ghosh2024octo} supports finetuning to new observation and action
spaces, while OpenVLA~\cite{kim2024openvla} uses LoRA for parameter-efficient
adaptation. OpenVLA-OFT~\cite{kim2025openvlaoft} improves
finetuning for inference speed and control performance.
Demonstrations also guide adaptation through policy context:
RICL~\cite{sridhar2025ricl} trains pretrained vision-language-action models
to adapt from retrieved demonstration segments without further parameter
updates. HyperVLA~\cite{xiong2025hypervla} generates compact
task-specific policies from language and an initial image.
WIZARD~\cite{bianchi2026wizard} predicts LoRA parameters for a frozen policy
from language and demonstration video, using fitted source-task adapters as
training targets. Our robotic extension generates per-image proposals from
demonstration images alone, averages completed updates into one
adapter, and learns from source query actions.

%% file: sections/method.tex
\section{Method}
\label{sec:method}

\begin{figure*}[t]
  \centering
  \resizebox{\textwidth}{!}{\input{figures/proposal2adapter_overview_vector}}
  \caption{\textbf{Proposal2Adapter with decode--aggregate overview.}
  A frozen, adapter-disabled encoder maps each unlabeled support image to an
  embedding. A learned conditioner and slot-specific factor decoders generate
  one low-rank update proposal per support example and adapter slot. The paired
  matrix products are formed before they are averaged, producing a single
  target-domain adapter shared by all queries. During episodic source training,
  disjoint query supervision updates only the hypernetwork; at deployment,
  adapter synthesis requires one forward pass and no gradients. Aggregate--decode instead averages the support codes before factor decoding.}
  \label{fig:proposal2adapter-overview}
\end{figure*}
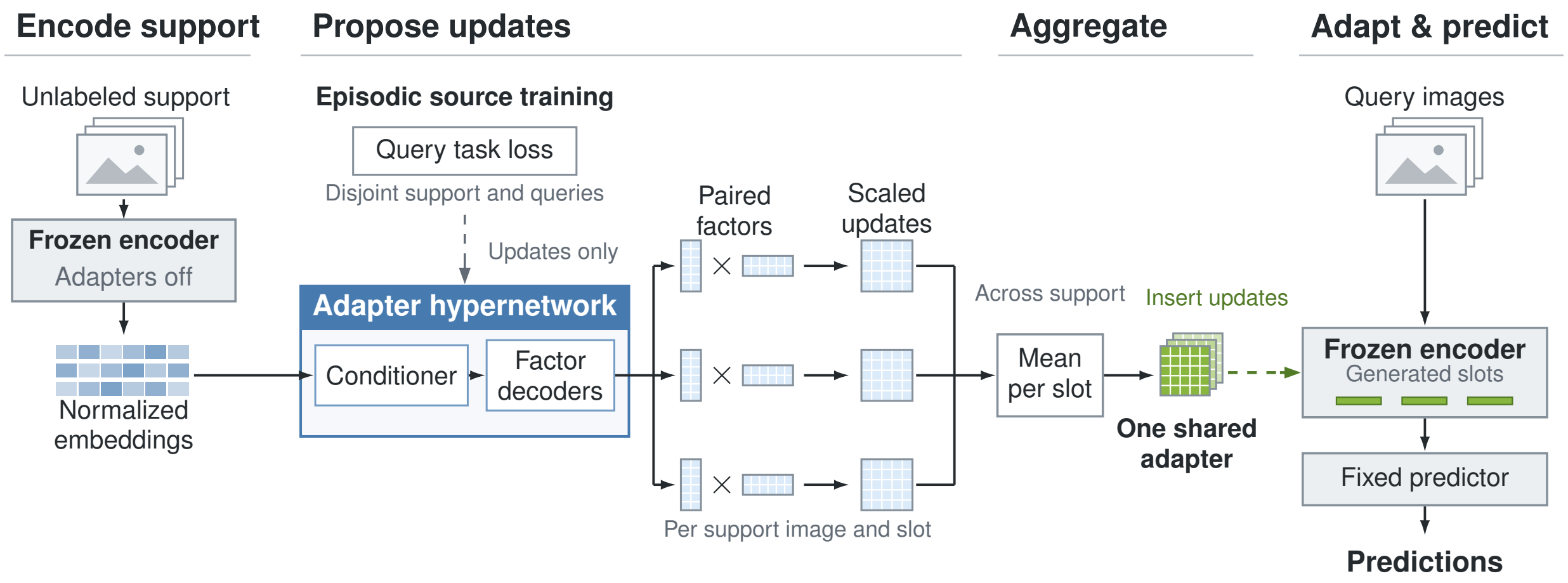

Given a small unlabeled support set from a target domain, Proposal2Adapter synthesizes a single adapter used to process queries from that domain.
Figure~\ref{fig:proposal2adapter-overview} depicts the canonical
decode--aggregate form. Both synthesis rules first encode each support image
with adapters disabled and map it to an instance code
(Secs.~\ref{sec:method-setup}--\ref{sec:method-support-coding}). A bank of slot-specific factor decoders maps learned support codes to low-rank weight updates
(Sec.~\ref{sec:method-hypernetwork}). Decode--aggregate applies this decoder
to every instance code before averaging the completed updates
(Sec.~\ref{sec:method-decode-aggregate}); aggregate--decode first averages the
codes and then decodes one update (Sec.~\ref{sec:method-aggregate-decode}).
Both rules yield one adapter shared by all target-domain queries and require no
optimization at inference time (Sec.~\ref{sec:method-training}).
\subsection{Problem Setting}
\label{sec:method-setup}

For a target domain $d$, we observe a small unlabeled support set
$\mathcal S_d=\{x_i\}_{i=1}^{S}$ and use it to adapt a model for queries
from the same domain. For image prediction, we use a frozen image encoder
$f_0$ with a frozen task predictor $g$. We adapt the encoder through additive updates to
$L$ selected linear weight matrices. Each selected matrix defines an
\emph{adapter slot} $\ell$, with frozen weights
$W_\ell\in\mathbb R^{d_\ell^{\mathrm{out}}\times d_\ell^{\mathrm{in}}}$,
update rank $r_\ell$, and scaling factor $\alpha_\ell$. For aggregation rule
$a\in\{\mathrm{D\to A},\mathrm{A\to D}\}$, we map $\mathcal S_d$ to a
\emph{target-domain adapter}
$\mathcal A_d^{(a)}=\{\bar M_\ell^{(a)}\}_{\ell=1}^{L}$, where
$\bar M_\ell^{(a)}$ is the synthesized update for slot $\ell$. All queries
from $d$ share this adapter. Domain identities serve only to group source
episodes.

\subsection{Support Embeddings and Instance Codes}
\label{sec:method-support-coding}

Let \(f_0\) denote the frozen base encoder evaluated without a generated adapter.
We encode each support image $x_i$ with the frozen encoder, producing a normalized embedding $z_i$. A shared
\emph{support conditioner} $C_\psi$ then maps $z_i$ to an instance code
$c_i\in\mathbb R^p$ for adapter synthesis,
\begin{equation}
  z_i=\operatorname{norm}\!\left(f_0(x_i)\right),
  \qquad
  c_i=C_\psi(z_i).
  \label{eq:support-code}
\end{equation}
Disabling adapters ensures that the features used to generate an adapter do
not depend on that adapter. Because the encoder is frozen, support embeddings
can also be cached. In our implementation,
$C_\psi=F_c\circ P_z$, where $P_z$ is a normalized linear projection and
$F_c$ is a two-layer MLP applied independently, with shared weights, to every support example. Layer widths and other implementation details are given in the supplement.
Layer dimensions and implementation
details are given in the supplement.

\subsection{Slot-Specific Low-Rank Update Decoder}
\label{sec:method-hypernetwork}

For any code $c\in\mathbb R^p$ and adapter slot $\ell$, two separate learned
\emph{factor decoders} produce the paired matrices
\begin{equation}
  U_\ell(c)=D_{\ell,U}(c), \qquad
  V_\ell(c)=D_{\ell,V}(c),
  \label{eq:slot-factors}
\end{equation}
where $U_\ell(c)\in\mathbb R^{d_\ell^{\mathrm{out}}\times r_\ell}$ and
$V_\ell(c)\in\mathbb R^{d_\ell^{\mathrm{in}}\times r_\ell}$. Their scaled
product defines a code-conditioned low-rank update,
\begin{equation}
  M_\ell(c)
  =
  \frac{\alpha_\ell}{r_\ell}
  U_\ell(c)V_\ell(c)^{\top}.
  \label{eq:code-conditioned-update}
\end{equation}
This gives an additive LoRA-style update~\cite{hu2022lora} of rank at most
$r_\ell$, with factors generated from a support-derived code rather than
stored as source-domain parameters. The support conditioner and all slot
factor decoders together form the adapter hypernetwork
$H_\Theta=(C_\psi,\{D_{\ell,U},D_{\ell,V}\}_{\ell=1}^{L})$. Composing
bias-free linear maps makes each factor branch linear in $c$ at deterministic
inference.

\subsection{Decode--Aggregate Synthesis}
\label{sec:method-aggregation}
\label{sec:method-decode-aggregate}

The canonical decode--aggregate rule evaluates the shared decoder for every
instance code. We use the shorthand
\begin{equation}
  U_{i\ell}=U_\ell(c_i), \qquad
  V_{i\ell}=V_\ell(c_i), \qquad
  M_{i\ell}=M_\ell(c_i),
  \label{eq:instance-proposal}
\end{equation}
and call $M_{i\ell}$ the instance-conditioned low-rank update proposal. For each slot, the proposals are averaged to form the target-domain adapter,
\begin{equation}
  \bar M_\ell^{(\mathrm{D\to A})}
  =
  \frac{1}{S}\sum_{i=1}^{S}M_{i\ell}.
  \label{eq:support-aggregation}
\end{equation}
The same average can be written as a product of factor matrices concatenated
across support examples,
\begin{equation}
  \bar M_\ell^{(\mathrm{D\to A})}
  =
  \frac{\alpha_\ell}{Sr_\ell}
  \bigl[U_{1\ell}\;\cdots\;U_{S\ell}\bigr]
  \bigl[V_{1\ell}\;\cdots\;V_{S\ell}\bigr]^{\top}.
  \label{eq:concatenated-aggregation}
\end{equation}
Although each proposal has rank at most $r_\ell$, their average can have
higher rank. Section~\ref{sec:aggregation-analysis} analyzes this capacity and
the support statistics retained by decode--aggregate.

\subsection{Aggregate--Decode Synthesis}
\label{sec:method-aggregate-decode}

The aggregate--decode rule first averages the instance codes produced by the
nonlinear support conditioner,
\begin{equation}
  \bar c=\frac{1}{S}\sum_{i=1}^{S}c_i,
  \label{eq:mean-support-code}
\end{equation}
and then evaluates each pair of factor decoders once,
\begin{equation}
  \begin{aligned}
  \bar M_\ell^{(\mathrm{A\to D})}
  &=M_\ell(\bar c) \\
  &=\frac{\alpha_\ell}{r_\ell}
    D_{\ell,U}(\bar c)D_{\ell,V}(\bar c)^{\top}.
  \end{aligned}
  \label{eq:aggregate-decode-update}
\end{equation}
Thus aggregate--decode produces a rank $r_\ell$ update and requires
one factor-decoder evaluation per slot rather than one per support code. With
the bias-free linear decoders used here,
\begin{equation}
  D_{\ell,U}(\bar c)=\frac{1}{S}\sum_i U_{i\ell}=\bar U_\ell,
  \;
  D_{\ell,V}(\bar c)=\frac{1}{S}\sum_i V_{i\ell}=\bar V_\ell,
  \label{eq:factor-means}
\end{equation}
so Eq.~\eqref{eq:aggregate-decode-update} is equivalently
$(\alpha_\ell/r_\ell)\bar U_\ell\bar V_\ell^{\top}$. The distinction between
this product of factor means and the mean of paired products in
Eq.~\eqref{eq:support-aggregation} is analyzed in
Section~\ref{sec:aggregation-analysis}.

\subsection{Training and Gradient-Free Adaptation}
\label{sec:method-training}

Under either rule $a\in\{\mathrm{D\to A},\mathrm{A\to D}\}$, the synthesized
slot updates define
\begin{equation}
  \mathcal A_d^{(a)}
  =\{\bar M_\ell^{(a)}\}_{\ell=1}^{L},
  \qquad
  \widetilde W_{\ell,d}^{(a)}=W_\ell+\bar M_\ell^{(a)}.
  \label{eq:variant-adapter}
\end{equation}
Both rules are invariant to support order and share the same conditioner and
factor-decoder architecture, dimensionalities, and parameter count. 

We train each hypernetwork using episodes sampled from labeled source domains. Each episode contains a support set $\mathcal S_d$ and a disjoint labeled
query set $\mathcal Q_d=\{(x_j^q,y_j^q)\}_{j=1}^{Q}$ from the same domain. We
generate $\mathcal A_d^{(a)}$ and evaluate the adapted model on the
queries. Writing $\Theta^{(a)}$ for the independently optimized hypernetwork
parameters, the query labels supervise
\begin{equation}
  \mathcal L^{(a)}(\Theta^{(a)})
  =
  \frac{1}{Q}\sum_{j=1}^{Q}
  \ell_{\mathrm{task}}\!\left(
    g\!\left(f_{\mathcal A_d^{(a)}}(x_j^q)\right),y_j^q
  \right).
  \label{eq:episodic-objective}
\end{equation}
The task loss can be chosen for classification, regression, or control.
Training updates only the support conditioner and slot factor decoders; the
base encoder and task predictor remain frozen. The hypernetwork is conditioned
only on support embeddings, without support labels, domain identifiers,
ordering information, or query images. At inference time, the selected
synthesis rule generates one adapter from a new domain's unlabeled support
set, and that adapter is shared by every query from the domain. Adaptation
requires no target labels, gradients, optimizer state, or stored source-domain
adapters.

%% file: figures/proposal2adapter_overview_vector.tex
\definecolor{overviewInk}{HTML}{252A2E}
\definecolor{overviewMuted}{HTML}{626B75}
\definecolor{overviewBorder}{HTML}{85909A}
\definecolor{overviewFrozen}{HTML}{ECEFF2}
\definecolor{overviewBlue}{HTML}{477BB0}
\definecolor{overviewBlueLight}{HTML}{DCEBFA}
\definecolor{overviewGreen}{HTML}{88B63F}
\definecolor{overviewGreenDark}{HTML}{527B28}

\begingroup
\newcommand{\overviewMatrix}[7]{%
  \begin{scope}[shift={(#1,#2)}]
    \fill[#7] (0,0) rectangle (#3,#4);
    \pgfmathtruncatemacro{\lastcolumn}{#5-1}
    \pgfmathtruncatemacro{\lastrow}{#6-1}
    \foreach \columnindex in {1,...,\lastcolumn} {
      \draw[white,line width=0.45pt]
        ({\columnindex*#3/#5},0) -- ({\columnindex*#3/#5},#4);
    }
    \foreach \rowindex in {1,...,\lastrow} {
      \draw[white,line width=0.45pt]
        (0,{\rowindex*#4/#6}) -- (#3,{\rowindex*#4/#6});
    }
    \draw[overviewBorder,line width=0.6pt] (0,0) rectangle (#3,#4);
  \end{scope}%
}
\newcommand{\overviewThumbnail}[2]{%
  \begin{scope}[shift={(#1,#2)}]
    \draw[draw=overviewBorder,fill=white,line width=0.6pt]
      (0.18,0.18) rectangle (1.16,0.84);
    \draw[draw=overviewBorder,fill=white,line width=0.6pt]
      (0.09,0.09) rectangle (1.07,0.75);
    \draw[draw=overviewBorder,fill=overviewFrozen!45,line width=0.65pt]
      (0,0) rectangle (0.98,0.66);
    \fill[overviewBorder!60] (0.09,0.12) -- (0.34,0.41) --
      (0.53,0.23) -- (0.66,0.33) -- (0.89,0.12) -- cycle;
    \fill[overviewBorder!80] (0.68,0.49) circle (0.055);
  \end{scope}%
}

\begin{tikzpicture}[
  x=1cm,y=1cm,
  font=\sffamily\fontsize{8}{9.5}\selectfont,
  text=overviewInk,
  flow/.style={-{Latex[length=1.7mm,width=1.1mm]},draw=overviewInk,line width=0.75pt},
  stage/.style={anchor=west,font=\sffamily\bfseries\fontsize{10}{11}\selectfont},
  note/.style={text=overviewMuted,font=\sffamily\fontsize{7}{8.5}\selectfont,align=center},
  frozen/.style={draw=overviewBorder,fill=overviewFrozen,line width=0.7pt,align=center},
  parameter/.style={draw=overviewGreenDark,dashed,line width=0.9pt,line cap=butt}
]
\useasboundingbox (0,0) rectangle (17.4,6.45);

% Four stages, matching the supplied overview.
\node[stage] at (0.15,6.12) {Encode support};
\node[stage] at (3.40,6.12) {Propose updates};
\node[stage] at (11.05,6.12) {Aggregate};
\node[stage] at (14.35,6.12) {Adapt \& predict};
\foreach \stageleft/\stageright in {0.15/2.85,3.40/10.65,11.05/13.85,14.35/17.25} {
  \draw[overviewBorder!60,line width=0.5pt]
    (\stageleft,5.83) -- (\stageright,5.83);
}

% Encode the unlabeled support with adapters disabled.
\node at (1.48,5.35) {Unlabeled support};
\overviewThumbnail{0.95}{4.30}
\draw[flow] (1.46,4.25) -- (1.46,4.02);
\node[frozen,minimum width=2.45cm,minimum height=0.83cm]
  (supportencoder) at (1.46,3.58)
  {\textbf{Frozen encoder}\\[2pt]\textcolor{overviewMuted}{Adapters off}};
\draw[flow] (supportencoder.south) -- (1.46,2.75);
\foreach \rowindex in {0,1,2} {
  \foreach \columnindex in {0,1,2,3,4,5} {
    \pgfmathtruncatemacro{\cellshade}{30+mod(\rowindex*3+\columnindex*2,5)*10}
    \fill[overviewBlue!\cellshade!white]
      ({0.72+\columnindex*0.245},{2.10+\rowindex*0.20})
      rectangle ++(0.225,0.15);
  }
}
\node[align=center] at (1.46,1.75) {Normalized\\embeddings};

% Episodic query supervision trains only the hypernetwork.
\node[font=\sffamily\bfseries\fontsize{8}{9.5}\selectfont]
  at (5.20,5.35) {Episodic source training};
\node[draw=overviewBorder,line width=0.6pt,minimum width=2.45cm,
  minimum height=0.52cm] (queryloss) at (5.20,4.78) {Query task loss};
\node[note] at (5.20,4.30) {Disjoint support and queries};
\draw[draw=overviewMuted,dashed,line width=0.7pt,line cap=butt]
  (5.20,4.08) -- (5.20,3.50);
\draw[overviewMuted,line width=0.7pt] (5.20,3.56) -- (5.20,3.49);
\fill[overviewMuted] (5.135,3.50) -- (5.265,3.50) -- (5.20,3.32) -- cycle;
\node[note,anchor=west] at (5.33,3.66) {Updates only};

% Shared conditioner and factor decoders.
\draw[draw=overviewBlue,fill=overviewBlue!5,line width=0.8pt]
  (3.40,1.65) rectangle (7.00,3.30);
\fill[overviewBlue] (3.40,2.82) rectangle (7.00,3.30);
\node[text=white,font=\sffamily\bfseries\fontsize{9}{10}\selectfont]
  at (5.20,3.06) {Adapter hypernetwork};
\node[draw=overviewBlue!75,fill=white,line width=0.6pt,
  minimum width=1.48cm,minimum height=0.66cm,anchor=west]
  (conditioner) at (3.55,2.32) {Conditioner};
\node[draw=overviewBlue!75,fill=white,line width=0.6pt,
  minimum width=1.40cm,minimum height=0.66cm,align=center,anchor=east]
  (decoders) at (6.85,2.32) {Factor\\decoders};
\draw[flow] (2.23,2.32) -- (conditioner.west);
\draw[flow] (conditioner.east) -- (decoders.west);

% Complete paired factor products for individual images and slots.
\node[align=center] at (8.16,4.13) {Paired\\factors};
\node[align=center] at (9.83,4.13) {Scaled\\updates};
\draw[overviewInk,line width=0.75pt] (decoders.east) -- (7.27,2.32);
\draw[overviewInk,line width=0.75pt] (7.27,1.12) -- (7.27,3.52);
\foreach \proposalcenter in {1.12,2.32,3.52} {
  \draw[flow] (7.27,\proposalcenter) -- (7.52,\proposalcenter);
  \overviewMatrix{7.57}{\proposalcenter-0.28}{0.22}{0.56}{2}{6}{overviewBlueLight}
  \node[font=\sffamily\bfseries\fontsize{11}{11}\selectfont]
    at (8.02,\proposalcenter) {$\times$};
  \overviewMatrix{8.25}{\proposalcenter-0.11}{0.55}{0.22}{6}{2}{overviewBlueLight}
  \draw[flow] (8.92,\proposalcenter) -- (9.42,\proposalcenter);
  \overviewMatrix{9.55}{\proposalcenter-0.28}{0.56}{0.56}{5}{5}{overviewBlueLight}
  \draw[overviewInk,line width=0.75pt]
    (10.15,\proposalcenter) -- (10.57,\proposalcenter);
}
\draw[overviewInk,line width=0.75pt] (10.57,1.12) -- (10.57,3.52);
\node[note] at (8.85,0.61) {Per support image and slot};

% Average completed updates across support examples, separately for each slot.
\node[note] at (11.62,3.20) {Across support};
\node[draw=overviewBorder,line width=0.7pt,minimum width=1.06cm,
  minimum height=0.90cm,align=center] (average) at (11.62,2.32) {Mean\\per slot};
\draw[flow] (10.57,2.32) -- (average.west);
\overviewMatrix{12.97}{2.24}{0.54}{0.54}{5}{5}{overviewGreen!35}
\overviewMatrix{12.90}{2.17}{0.54}{0.54}{5}{5}{overviewGreen!55}
\overviewMatrix{12.83}{2.10}{0.54}{0.54}{5}{5}{overviewGreen}
\draw[flow] (average.east) -- (12.77,2.32);
\node[font=\sffamily\bfseries\fontsize{8}{9.5}\selectfont,align=center]
  at (13.12,1.55) {One shared\\adapter};
\node[note,text=overviewGreenDark] at (13.45,3.15) {Insert updates};

% Apply the shared adapter to queries, with the fixed predictor unchanged.
\node at (15.73,5.35) {Query images};
\overviewThumbnail{15.20}{4.30}
\node[frozen,minimum width=2.68cm,minimum height=0.98cm]
  (queryencoder) at (15.73,2.35) {};
\node[font=\sffamily\bfseries\fontsize{8.5}{10}\selectfont]
  at (15.73,2.61) {Frozen encoder};
\node[note] at (15.73,2.34) {Generated slots};
\foreach \slotleft in {14.76,15.48,16.20} {
  \draw[draw=overviewGreenDark,fill=overviewGreen,line width=0.45pt]
    (\slotleft,2.00) rectangle ++(0.49,0.08);
}
\draw[flow] (15.73,4.25) -- (queryencoder.north);
\draw[parameter] (13.57,2.35) -- (14.19,2.35);
\draw[overviewGreenDark,line width=0.9pt] (14.14,2.35) -- (14.20,2.35);
\fill[overviewGreenDark] (14.19,2.285) -- (14.19,2.415) -- (14.37,2.35) -- cycle;
\node[frozen,minimum width=2.68cm,minimum height=0.57cm]
  (predictor) at (15.73,1.18) {Fixed predictor};
\draw[flow] (queryencoder.south) -- (predictor.north);
\draw[flow] (predictor.south) -- (15.73,0.56);
\node[font=\sffamily\bfseries\fontsize{9}{10}\selectfont]
  at (15.73,0.28) {Predictions};
\end{tikzpicture}
\endgroup

%% file: sections/analysis.tex
\section{Analysis of Aggregation Order}
\label{sec:aggregation-analysis}

We compare the support statistics and rank capacity retained by the two
synthesis rules. We consider the bias-free linear factor decoders used at
deterministic inference (Sec.~\ref{sec:method-hypernetwork}); the support
conditioner may be nonlinear. We hold these learned maps fixed when comparing
the operations below; the reported variants are trained independently. Full
proofs are given in the supplement.

\paragraph{Statistics retained by aggregation order.}
Because each factor is linear in $c_i$, every entry of $M_\ell(c_i)$ is a
quadratic form in that code. Thus, there exists a learned linear map
$\mathcal L_\ell$ such that decode--aggregate depends on the empirical second
moment of the support codes,
\begin{equation}
  \bar M_\ell^{(\mathrm{D\to A})}
  =
  \mathcal L_\ell(G_{\mathcal S}),
  \qquad
  G_{\mathcal S}
  =
  \frac{1}{S}\sum_{i=1}^{S}c_ic_i^\top.
  \label{eq:second-moment-map}
\end{equation}
Here $\mathcal L_\ell$ is a learned linear map from $p\times p$ matrices to
updates with the same shape as $W_\ell$.

To separate the mean code from variation among support examples, define
\begin{equation}
  \bar c=\frac{1}{S}\sum_i c_i,
  \qquad
  \Sigma_{\mathcal S}
  =
  \frac{1}{S}\sum_i(c_i-\bar c)(c_i-\bar c)^\top.
\end{equation}
Since $G_{\mathcal S}=\bar c\bar c^\top+\Sigma_{\mathcal S}$ and
$M_\ell(\bar c)=\mathcal L_\ell(\bar c\bar c^\top)$, the two synthesis rules
are related exactly by
\begin{equation}
  \bar M_\ell^{(\mathrm{D\to A})}
  =
  \bar M_\ell^{(\mathrm{A\to D})}
  +
  \mathcal L_\ell(\Sigma_{\mathcal S}).
  \label{eq:mean-covariance-decomposition}
\end{equation}
% This identity characterizes the architectural difference between the two operations, rather than a numerical relation between independently trained
% checkpoints.
Decode--aggregate can therefore use variation among support codes, whereas
aggregate--decode depends on their mean. This comparison is in the learned
code space: because $c_i=C_\psi(z_i)$ and $C_\psi$ is nonlinear, the mean code
can itself encode richer distributional statistics than the mean input
embedding. The support-variation term can also vanish under the learned map.

% Schedule the benchmark float before the next page starts.
\begin{table*}[t]
  \centering
  \caption{Comparison on the WILDS OOD-test splits. D--A denotes
  decode--aggregate and A--D denotes aggregate--decode. Proposal2Adapter and
  our support-independent ERM and LoRA controls use five seeds on every
  dataset; the controls use no target support. All iWildCam rows use only the
  original labeled source pool. Parentheses give standard deviations, and
  $\dagger$ marks each dataset's official metric.
  Within each backbone group (CNN, ViT-B/16, and ViT-L/14), the best value
  among the methods shown is bold; ties at the reported precision are all bold.}
  \label{tab:wilds-main-comparison}
  \setlength{\tabcolsep}{2.9pt}
  \renewcommand{\arraystretch}{1.08}
  \small
  \begin{tabular}{llcccccc}
    \toprule
    & & \multicolumn{2}{c}{iWildCam} & \multicolumn{2}{c}{FMoW} &
    \multicolumn{2}{c}{PovertyMap} \\
    \cmidrule(lr){3-4}\cmidrule(lr){5-6}\cmidrule(lr){7-8}
    Method & Backbone & Macro-F1$^{\dagger}$ & Acc. & WC Acc.$^{\dagger}$ &
    Acc. & WC Pearson $r^{\dagger}$ & Pearson $r$ \\
    \midrule
    ARM-CML~\cite{zhang2021arm} & CNN & 28.6 (0.1) & 70.5 (0.6) & 27.2 (0.38) & 45.7 (0.28) & 0.37 (0.08) & 0.75 (0.04) \\
    MABN~\cite{wu2024mabn} & CNN & \textbf{38.3 (1.2)} & \textbf{78.4 (0.6)} & \textbf{36.6 (0.41)} & \textbf{53.2 (0.52)} & \textbf{0.56 (0.05)} & \textbf{0.84 (0.04)} \\
    \midrule
    ERM; frozen prompt head & ViT-B/16 & 35.21 (1.17) & 75.30 (0.46) & 39.42 (1.08) & 57.25 (1.34) & 0.517 (0.022) & 0.839 (0.009) \\
    LoRA; frozen prompt head & ViT-B/16 & 31.83 (1.04) & 73.24 (0.57) & 39.50 (1.04) & 53.67 (0.60) & 0.518 (0.007) & 0.842 (0.004) \\
    L2C~\cite{chi2025l2c} & ViT-B/16 & 35.2 (0.3) & 73.4 (0.4) & 40.9 (0.4) & 54.8 (0.1) & 0.50 (0.02) & 0.80 (0.03) \\
    DA-MergeLoRA~\cite{reid2026damerge} & ViT-B/16 & 33.62 (1.63) & 73.70 (1.49) & \textbf{52.26 (0.16)} & 57.60 (0.14) & 0.56 (0.01) & 0.84 (0.01) \\
    Proposal2Adapter (D--A) & ViT-B/16 & 37.82 (1.67) & 74.91 (1.59) & 43.94 (2.59) & 57.45 (1.20) & \textbf{0.580 (0.035)} & \textbf{0.860 (0.008)} \\
    Proposal2Adapter (A--D) & ViT-B/16 & \textbf{39.22 (0.84)} & \textbf{75.36 (1.10)} & 46.48 (2.65) & \textbf{58.95 (1.16)} & 0.562 (0.022) & 0.857 (0.008) \\
    \midrule
    L2C~\cite{chi2025l2c} & ViT-L/14 & \textbf{48.6 (0.4)} & 77.3 (0.1) & \textbf{46.7 (0.3)} & \textbf{61.4 (0.3)} & \textbf{0.56 (0.02)} & \textbf{0.84 (0.03)} \\
    Proposal2Adapter (D--A) & ViT-L/14 & 44.85 (1.37) & \textbf{77.45 (0.67)} & -- & -- & -- & -- \\
    \bottomrule
  \end{tabular}
\end{table*}

\paragraph{Rank capacity and inductive bias.}
Aggregate--decode always produces
\begin{equation}
  \operatorname{rank}
  \bigl(\bar M_\ell^{(\mathrm{A\to D})}\bigr)
  \le r_\ell.
  \label{eq:aggregate-decode-rank-bound}
\end{equation}
Decode--aggregate can instead combine the directions of multiple rank-$r_\ell$
proposals. Let
$m_\ell=\min(d_\ell^{\mathrm{out}},d_\ell^{\mathrm{in}})$. Its update satisfies
\begin{align}
  \operatorname{rank}
  \bigl(\bar M_\ell^{(\mathrm{D\to A})}\bigr)
  &\le
  \min\!\left\{
    m_\ell,\,
    r_\ell\operatorname{rank}(G_{\mathcal S})
  \right\}
  \notag\\
  &\le
  \min\!\left\{
    m_\ell,\,
    r_\ell\min(S,p)
  \right\}.
  \label{eq:aggregate-rank-bound}
\end{align}
Rank $Sr_\ell$ is attainable when the support codes are linearly independent,
the scale is nonzero, and the slot dimensions and decoder capacity are
sufficiently large. The supplement gives the width-dependent bound and attainability conditions.
The bound describes the rank the architecture can represent.  Decode--aggregate thus retains an explicit
support-variation term and has greater potential rank capacity.
Aggregate--decode instead imposes a fixed-rank bottleneck, evaluates each
factor decoder only once per slot, and may provide useful regularization.
% These are distinct inductive biases: the capacity result does not imply that
% decode--aggregate must yield better predictions.

%% file: sections/experiments.tex
\section{Experiments}
\label{sec:experiments}

% Schedule the source-scaling figure near its results.
\begin{figure*}[t]
  \centering
  \includegraphics[width=0.98\textwidth]{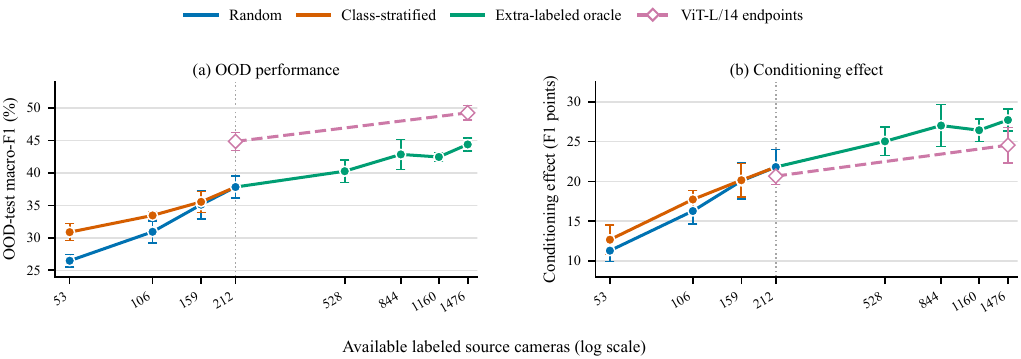}
  \caption{Source-camera scaling for Proposal2Adapter with decode--aggregate
  synthesis. (a) OOD-test macro-F1.
  (b) Conditioning effect, measured as correct-camera minus matched
  wrong-camera macro-F1 on a fixed 44-camera control cohort. Error bars are
  sample standard deviations over five seeds. The ViT-L/14 trace contains only
  the labeled-only and full +1,264-extra-camera endpoints; all other traces use
  ViT-B/16. Extra-camera coordinates count original plus added cameras. The shared 212-camera
  endpoint is drawn once in the Random color.}
  \label{fig:iwildcam-camera-ladders}
\end{figure*}

\paragraph{Evaluation overview.}
We evaluate Proposal2Adapter in two settings. On WILDS~\cite{koh2021wilds},
we compare against existing few-shot test-time domain adaptation methods and
study the effects of aggregation order, source-domain supervision, and target
support size
(Sections~\ref{sec:experiments-main-comparison}--\ref{sec:experiments-support-size}).
We then extend the evaluation to robot control on LIBERO~\cite{liu2023libero},
using visual demonstrations to adapt a frozen policy to unseen tasks. This
study compares adaptation methods across four held-out suites and examines
the effect of the number of support demonstrations
(Section~\ref{sec:experiments-libero}).

For the WILDS support controls, we define the conditioning effect for
evaluation metric $M$ as
\begin{equation}
  \Delta_{\mathrm{cond}}=M_{\mathrm{correct}}-M_{\mathrm{wrong}}.
  \label{eq:experimental-evidence}
\end{equation}
The same queries are evaluated using adapters generated from correct- or
wrong-domain support. A positive gap establishes functional support dependence,
not recovery of a unique causal domain mechanism. We use official WILDS metrics
for primary comparisons. Experiments using normally hidden labels are identified
as oracle diagnostics and are excluded from the standard benchmark.

\subsection{Benchmark Comparisons on WILDS}
\label{sec:experiments-main-comparison}

We evaluate Proposal2Adapter on iWildCam~\cite{beery2021iwildcam},
FMoW~\cite{christie2018fmow}, and PovertyMap~\cite{yeh2020povertymap} using
the official WILDS~\cite{koh2021wilds} splits and metrics; following prior
FSTT-DA work, PovertyMap uses RGB3 fold A. The training sets
contain 212 eligible camera domains for iWildCam; for FMoW and PovertyMap, we
use the fine-grained definitions, giving 55 region--year and 26
country--urban/rural training domains, respectively.
Table~\ref{tab:wilds-main-comparison} groups methods by backbone and reports
published baseline results~\cite{reid2026damerge,chi2025l2c} alongside our
results.
%The prior-method results in Table~\ref{tab:wilds-main-comparison} are transcribed from Table~1b of DA-MergeLoRA~\cite{reid2026damerge}.
%% this can be moved to supplemental material
%Our principal row uses the Proposal2Adapter support conditioner and a frozen
%CLIP ViT-B/16 backbone. The iWildCam result uses five seeds and only the
%original labeled training split: it uses neither extra-unlabeled cameras,
%self-supervision on the extra pool, nor oracle labels. FMoW and PovertyMap use
%five seeds. We additionally report a ViT-L/14 Proposal2Adapter result for
%iWildCam. All results are means with sample standard deviations in parentheses.
To anchor the comparison, we train two support-independent source baselines:
full visual ERM and a rank-16 LoRA, both with frozen task heads: CLIP prompt
classifiers for iWildCam and FMoW, and a fitted MLP regressor for PovertyMap.
Each is independently calibrated by a validation-only learning-rate sweep
before locked OOD-test evaluation; details are given in the supplement. Our
ERM reaches 35.21 macro-F1 on iWildCam, nearly identical to the 35.0 reported
for standard full fine-tuning of OpenAI CLIP ViT-B/16 by
Goyal et al.~\cite{goyal2023flyp}.

Our strongest results are on iWildCam. With ViT-B/16, aggregate--decode reaches
$39.22\pm0.84$ macro-F1, the highest reported mean among the same-backbone
methods shown and 4.02 points above L2C. Decode--aggregate reaches
$37.82\pm1.67$, 2.62 points above L2C. Our ViT-L/14 result uses
decode--aggregate and further improves over its ViT-B/16 counterpart, although
L2C retains higher macro-F1 with that backbone. On FMoW, aggregate--decode
reaches 46.48 worst-region accuracy and 58.95 overall accuracy; DA-MergeLoRA is
higher on the official metric, while Proposal2Adapter is higher overall. On
PovertyMap, decode--aggregate reaches .580 worst-group and .860 overall Pearson
correlation, the highest means among the compared methods.

%All methods use the official WILDS data splits and metrics, but the comparison
%is not a parameter-matched architecture ablation. In particular, the methods
%construct source domains differently: Proposal2Adapter uses camera domains for
%iWildCam, region-year domains for FMoW, and country-urban domains for
%PovertyMap. We therefore interpret Table~\ref{tab:wilds-main-comparison} as a
%benchmark comparison and study the role of source-domain availability
%separately in the source-domain scaling experiments below.

\subsection{Effect of Aggregation Order}
\label{sec:experiments-aggregation-ablation}

The two variants have identical trainable architectures and parameter counts and are
trained independently with the same recipe. We test whether the additional
capacity analyzed in Section~\ref{sec:aggregation-analysis} translates into
better predictions.
We evaluate both orders under two source-supervision settings. The standard
setting uses the original labeled source pool. The oracle-supervision
diagnostic retains that pool and adds the 1,264 cameras from the normally
unlabeled iWildCam release having at least 32 usable oracle-labeled images,
sufficient for sample-disjoint $S=Q=16$ episodes. Their hidden labels
are used only for episodic source training; cohort construction is detailed in
the supplement.

\begin{table*}[t]
  \centering
  \caption{Aggregation-order comparison on the locked iWildCam OOD test.
  Parentheses give sample standard deviations over five matched seeds.
  $\Delta_{\mathrm{cond}}$ is the correct-minus-wrong support macro-F1 gap.
  Rank statistics summarize test-support updates by averaging slots within
  cameras and cameras within each seed; $E_{>16}$ is the percentage of squared
  spectral energy beyond rank 16. ``+1,264 cameras'' retains the original
  source pool and adds every eligible extra camera using normally hidden labels.}
  \label{tab:aggregation-order}
  \setlength{\tabcolsep}{4.0pt}
  \renewcommand{\arraystretch}{1.08}
  \small
  \begin{tabular}{llrrrrrr}
    \toprule
    Source & Order & Macro-F1 & Acc. & $\Delta_{\mathrm{cond}}$ & Num. rank & Entropy rank & $E_{>16}$ (\%) \\
    \midrule
    Original labeled & D--A & 37.82 (1.67) & 74.91 (1.59) & 21.81 (2.19) & 148.38 (4.05) & 29.36 (1.39) & 3.05 (0.52) \\
    Original labeled & A--D & 39.22 (0.84) & 75.36 (1.10) & 23.54 (0.66) & 16.00 (0.00) & 15.06 (0.01) & $<10^{-12}$ \\
    +1,264 cameras & D--A & 44.39 (0.99) & 75.12 (3.15) & 27.72 (1.36) & 164.51 (4.17) & 33.24 (0.95) & 4.51 (0.48) \\
    +1,264 cameras & A--D & 43.56 (1.27) & 75.76 (1.48) & 28.25 (1.70) & 16.00 (0.00) & 15.10 (0.02) & $<10^{-12}$ \\
    \bottomrule
  \end{tabular}
\end{table*}

The aggregate--decode minus decode--aggregate macro-F1 contrast is 1.40 points
with the original labeled pool (descriptive paired 95\% interval
$[-0.70,3.49]$) and reverses to $-0.83$ with the 1,264 extra cameras
($[-3.13,1.47]$).
Table~\ref{tab:aggregation-order} also reports the realized update geometry
predicted by Section~\ref{sec:aggregation-analysis}. Entropy rank is
$\exp(-\sum_j p_j\log p_j)$, where $p_j=\sigma_j/\sum_k\sigma_k$, and measures
how many singular directions carry substantial spectral mass.
Decode--aggregate has mean entropy ranks of 29.36 and 33.24 under the original
and expanded source pools, compared with 15.06 and 15.10 for
aggregate--decode. It also places 3.05\% and 4.51\% of squared spectral energy
beyond rank 16.
Across the remote-sensing datasets in Table~\ref{tab:wilds-main-comparison},
aggregate--decode has higher FMoW means, whereas decode--aggregate has higher
PovertyMap means.

% SUPPLEMENTAL-ONLY MATERIAL (omit from the main paper):
% - Full per-seed results and secondary metrics (iWildCam macro recall, FMoW
%   macro-F1, and PovertyMap MSE).
% - Training recipes and protocol details: iWildCam uses the complete original
%   labeled set, meta-batch K=4, and no extra-unlabeled, oracle, or
%   self-supervised data; FMoW uses region-year domains; PovertyMap uses
%   country-urban domains; all three headline campaigns train for 20K steps.
% - Manifest, checkpoint, and locked-test provenance for each reported run.

\subsection{Effect of Source-Domain Supervision}
\label{sec:experiments-camera-ladders}

We next study whether richer source supervision improves prediction and makes
target support more useful. On iWildCam, each camera defines a domain.
These source-scaling experiments retain decode--aggregate as the canonical
synthesis rule.
Alongside overall performance, we measure the benefit of using support from
the query camera. We keep the queries fixed and replace matching support with
equally sized support from another camera. The resulting performance gap
measures how much matching the support helps.

With random selection, expanding the original training pool from 53 to 212
cameras improves macro-F1 by 11.35 points and widens the matching-support gap
(Fig.~\ref{fig:iwildcam-camera-ladders}). To examine whether missing classes
explain this trend, we also use class-stratified camera subsets that retain
all source classes. Preserving class coverage improves results for smaller
training pools, but performance continues to improve as more cameras are
added.

We then incrementally add subsets of the 1,264-camera oracle pool defined
above. Both decode--aggregate backbones improve, with ViT-L/14 reaching 49.25 macro-F1
(Fig.~\ref{fig:iwildcam-camera-ladders}).
The performance gains are driven primarily by better prediction with matching
target support. Across the ordinary and oracle ladders, ViT-B/16
wrong-support macro-F1 changes only from $16.37 \pm 0.55$ with 53 source
cameras to $17.58 \pm 1.03$ with 1,476 cameras, while correct-support
performance improves substantially. Camera count grows with the
amount and variety of training data, so this experiment measures the benefit
of richer source supervision.

Finally, in separate experiments, we use coarse and fine domain definitions to test the effect of
source-domain richness. The FMoW and PovertyMap results in
Table~\ref{tab:wilds-main-comparison} use the fine definitions: region and year
for FMoW, and country and urban/rural status for PovertyMap. Coarser grouping
gives lower held-out test scores on both datasets. In the FMoW experiment
(Supplementary Section~3.3), region-year grouping improves worst-region
accuracy and makes matching target support more useful than year-only
grouping. On PovertyMap, finer grouping improves the test score but lowers
validation performance. These comparisons support
our hypothesis that richer source variation improves adaptation, with
clearer evidence on FMoW. Limited source-domain richness may help explain
the remaining FMoW official-metric gap to DA-MergeLoRA.

% SUPPLEMENTAL-ONLY MATERIAL (omit from the main paper):
% - Exact nested-camera construction, seed-specific membership, image and
%   sequence counts, represented-class counts, and sampler-weighted class TV.
% - Full per-seed primary and support-control scores, paired adjacent-budget
%   contrasts, and the camera-bootstrap pass criteria for the controls.
% - The 20K-step K=4 training recipe, checkpoint selection, test-lock
%   provenance, and the distinction between the primary 48-camera cohort and
%   the matched 44-camera correct-versus-wrong control cohort.
% - Extra-pool label-space audit, taxonomy warnings, and the 2+2
%   ordinary/oracle episode mixture used only in the oracle ladder.

\begin{table*}[t]
  \centering
  \caption{LIBERO held-out-suite success (\%) with one visual support
  demonstration. Parentheses give sample standard deviations across three
  seeds. DA-MergeLoRA is evaluated using our implementation. The frozen policy achieves zero
  success in every suite; LIBERO-10 measures full-task completion.}
  \label{tab:libero-main}
  \setlength{\tabcolsep}{8pt}
  \small
  \begin{tabular}{lccc}
    \toprule
    Held-out suite & Conventional LoRA & DA-MergeLoRA & Proposal2Adapter (D--A) \\
    \midrule
    Spatial & 4.20 (1.22) & 0.07 (0.12) & \textbf{17.87 (2.32)} \\
    Object & 5.80 (0.92) & 0.00 (0.00) & \textbf{11.60 (6.13)} \\
    Goal & 7.73 (4.80) & 0.40 (0.69) & \textbf{8.67 (2.14)} \\
    LIBERO-10 & 0.00 (0.00) & 0.00 (0.00) & \textbf{0.33 (0.42)} \\
    \bottomrule
  \end{tabular}
\end{table*}

\subsection{Target Support Size}
\label{sec:experiments-support-size}

\begin{figure}[t]
  \centering
  \includegraphics[width=\columnwidth]{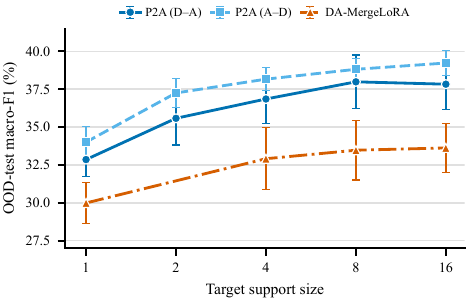}
  \caption{Sensitivity to target support size for decode--aggregate
  and aggregate--decode Proposal2Adapter on the iWildCam OOD test.
  Points show mean macro-F1 and error bars show sample standard deviation over
  five seeds for each Proposal2Adapter order and three seeds for DA-MergeLoRA. DA-MergeLoRA
  values are from its supplementary Appendix~A.3.3, Table~5; no size-2 result
  is reported.}
  \label{fig:iwildcam-support-size}
\end{figure}

We reduce target support at inference while keeping trained
checkpoints and query images fixed. Aggregate--decode has higher mean macro-F1
at every cap, but relative sensitivity is similar: decode--aggregate and
aggregate--decode retain 97.42\% and 97.29\% of their 16-shot performance at
four shots, and 86.86\% and 86.70\% at one. Their one-shot correct-minus-wrong
support gaps remain 17.18 and 18.23 points, respectively. As support decreases
from 16 to one image, decode--aggregate entropy rank contracts from 29.36 to
15.17, whereas aggregate--decode remains near 15.1; both updates have numerical
rank 16 at one shot. Both Proposal2Adapter orders achieve higher mean macro-F1 than
DA-MergeLoRA~\cite{reid2026damerge} at every support size reported for both
methods (Fig.~\ref{fig:iwildcam-support-size}). 
%The published supplementary results reflect its own training procedure.

% SUPPLEMENTAL-ONLY MATERIAL (omit from the main paper):
% - Full per-seed Proposal2Adapter scores, accuracy, macro recall, paired changes
%   from size 16, and realized support counts for short target cameras.
% - Proposal2Adapter uses fixed validation-selected checkpoints and nested support
%   manifests; only target inference support changes. The analysis was requested
%   post hoc and does not select a new checkpoint or operating support size.
% - Exact primary and support-excluded cohorts, correct-versus-wrong controls,
%   adapter-geometry diagnostics, and artifact/lock provenance.
% - DA-MergeLoRA source audit: the values are in the OpenReview supplementary
%   Appendix A.3.3, Table 5. GitHub commit
%   1b4a420a2c28cb0d149ebc078de9c51e113829b6 contains appendix scripts for
%   prediction examples, domain correlation, and t-SNE, but no support-size
%   result table or plotting script.

\subsection{Robot Policy Adaptation}
\label{sec:experiments-libero}

We evaluate adaptation of the frozen $\pi_{0.5}$ robot
policy~\cite{black2025pi05} on LIBERO~\cite{liu2023libero}.
We hold out Spatial, Object, Goal, and LIBERO-10 in turn, training on 30 tasks
from the other three suites and testing on the ten tasks in the held-out
suite. Each task defines an adaptation domain, and the entire target suite
is excluded from source training.

\begin{samepage}
We use decode--aggregate synthesis and insert the aggregated proposals into
selected language and action attention projections of the policy.
Source query actions provide the episodic training loss for the hypernetwork.
One demonstration of the target task supplies 16 images for adapter
generation. The conditioner receives no actions, states, or instructions;
the executing policy receives its instruction, live observations, and robot
state. We select checkpoints
on source validation and evaluate 50 official initial states per task.
Table~\ref{tab:libero-main} reports three-seed means and sample standard deviations.

\end{samepage}

Our approach achieves higher mean success than source-trained conventional LoRA
in all four suites, with the largest gains on Spatial and Object. Spatial
success rises from 4.20\% to 17.87\%, while full-task completion on LIBERO-10
remains near zero. Our DA-MergeLoRA implementation also achieves low success under this
protocol; the supplement reports implementation diagnostics.

We also vary target support from one to three, five, and ten demonstrations,
with 16 images per demonstration, while keeping the trained models fixed.
Spatial success reaches 19.73\% with five demonstrations, but more support
gives no consistent improvement across suites. The supplement reports the
full sweep and training details.

%% file: sections/limitations.tex
\section{Limitations}
\label{sec:limitations}

% TODO: Discuss label availability, domain coverage, compute, and protocol scope.

In Section~\ref{sec:aggregation-analysis}, we characterized the information
captured by proposal aggregation. With linear factor decoders,
decode--aggregate cannot distinguish support sets with the same empirical
second moment $G_{\mathcal S}$ in learned code space, while aggregate--decode
cannot distinguish sets with the same mean code. These restrictions apply
even when the examples themselves differ. They do
not imply that the adapter is quadratic in raw images or frozen encoder
embeddings, because the support conditioner may be nonlinear.

\begin{samepage}
In Section~\ref{sec:experiments-libero}, we evaluated robot control on LIBERO.
We were unable to reproduce the reported results of
WIZARD~\cite{bianchi2026wizard}. With runnable author code unavailable, we
implemented the method from its published description and obtained no
successful rollouts across the four suites (0/500 per suite). The result
reflects our reimplementation and does not establish the performance of the
authors' method.

\end{samepage}

%% file: sections/conclusion.tex
\section{Conclusion}
\label{sec:conclusion}

% TODO: Summarize the findings without introducing new claims.

We introduced Proposal2Adapter, a source-shared hypernetwork that synthesizes
a target-domain adapter from a small unlabeled support set without test-time
optimization. We studied two permutation-invariant synthesis rules with
distinct inductive biases. With linear factor decoders, decode--aggregate
retains variation among learned support codes and can combine
instance-conditioned low-rank proposals into higher-rank updates, whereas
aggregate--decode operates on the mean support code and remains rank at most
$r$. On standard iWildCam, aggregate--decode achieves 39.22 macro-F1 with
ViT-B/16, the strongest result among the compared same-backbone methods;
decode--aggregate reaches 37.82, and both orders retain approximately 97\% of
their respective 16-shot performance with four support images. Across
source-camera and domain-granularity studies,
decode--aggregate improves as source supervision becomes richer and aligned
with support-identifiable variation, with gains driven primarily by better
matched-support adaptation. Proposal2Adapter also extends to LIBERO, where one
visual demonstration raises Spatial success from 4.20\% for conventional LoRA
to 17.87\%. Together, these results establish source-shared support-to-adapter
generation as a scalable approach to few-shot test-time adaptation without
test-time gradients, stored source-domain adapters, or parameter growth with
the number of source domains.

%% file: sections/acknowledgments.tex
\section*{Acknowledgments}

The compute capacity at the Redtail AI Factory supercomputer from the Center for High Performance Computing at the University of Utah are gratefully acknowledged.

%% file: sections/llm_usage.tex
\section*{LLM Usage Statement}

We used OpenAI tools (ChatGPT and Codex) to assist with drafting and revising text
throughout the main paper and supplementary material, identifying related
work, and reviewing mathematical derivations. The tools also assisted with
writing and debugging experimental and plotting code and preparing figures.
The authors retain responsibility for the scientific claims, citations,
code, and reported results.

%% file: sections/supplementary_overview.tex
\section*{Key Additional Source-Richness Controls}

This opening section foregrounds additional controls that extend
main-paper Section~5.3 and Table~2.  The Proposal2Adapter rows retain the
submitted architecture, hyperparameters, target-support protocol, and
validation-only checkpoint selection; only the available labeled source
cameras change.  All entries below are five-seed means on the locked iWildCam
OOD test.  
``+1,264 extra'' uses labels normally hidden by the benchmark and
is therefore a diagnostic rather than part of the standard benchmark.

\begin{center}
  \centering
  \captionof{table}{Source expansion on iWildCam. Values are macro-F1 on the
  0--100 scale. Each method adds the same 1,264 eligible extra cameras to its
  own ordinary source pool (212 cameras for Proposal2Adapter and 243 for
  DA-MergeLoRA). Rel. is the gain relative to that method's ordinary result.}
  \label{tab:supp-key-source-richness}
  \setlength{\tabcolsep}{2.6pt}
  \scriptsize
  \begin{tabular}{@{}lrrrr@{}}
    \toprule
    Method / order & Ordinary & +1,264 & $\Delta$ & Rel. \\
    \midrule
    Proposal2Adapter D--A & 37.82 & 44.39 & +6.56 & +17.35\% \\
    Proposal2Adapter A--D & 39.22 & 43.56 & +4.33 & +11.05\% \\
    DA-MergeLoRA, $K=10$ & 33.25 & 35.89 & +2.64 & +7.94\% \\
    DA-MergeLoRA, $K=20$ & 32.57 & 32.94 & +0.38 & +1.17\% \\
    \bottomrule
  \end{tabular}
\end{center}

Both Proposal2Adapter orders improve in all five matched seeds after source
expansion.  The released-style $K=10$ DA-MergeLoRA control improves in only four of
five seeds. Thus, under
this diagnostic, Proposal2Adapter shows the larger observed benefit from the
additional camera supervision compared to DA-MergeLoRA. We further tested whether using a larger number of clusters ($K=20$) with DA-MergeLoRA would help utilize richer source availability; however, results deteriorated under both ordinary and oracle settings.

Correct-minus-wrong-support macro-F1 gap
also widens from 21.81 to 27.72 points for Proposal2Adapter decode--aggregate (D--A) and from
23.54 to 28.25 for aggregate--decode (A--D), compared with only 17.86 to 19.91 for
DA-MergeLoRA at $K=10$. Section~\ref{sec:supp-source-scale-iwildcam} describes in detail both full
camera ladders and the DA-MergeLoRA control. 

%% file: sections/supplementary_analysis.tex
\section{Proofs for Aggregation Order}
\label{sec:supp-aggregation-proofs}

We prove the properties stated in the analysis section of the main paper. For clarity, we
suppress the adapter-slot index. Let $c_i\in\mathbb R^p$, let
$U(c_i)\in\mathbb R^{o\times r}$ and
$V(c_i)\in\mathbb R^{d\times r}$, and let $s=\alpha/r$. At deterministic
inference, the decoder uses bias-free linear factor branches. For the
$q$th columns $u_q(c)$ and $v_q(c)$, there exist matrices
$A_q\in\mathbb R^{o\times p}$ and $B_q\in\mathbb R^{d\times p}$ such that
\begin{equation}
  u_q(c)=A_qc,
  \qquad
  v_q(c)=B_qc.
  \label{eq:supp-factor-column-maps}
\end{equation}
Define $M(c)=sU(c)V(c)^\top$ and
$\bar c=S^{-1}\sum_i c_i$. The two synthesis rules are
\begin{equation}
  \begin{aligned}
    \bar M^{(\mathrm{D\to A})}
    &=\frac{1}{S}\sum_{i=1}^{S}M(c_i), \\
    \bar M^{(\mathrm{A\to D})}
    &=M(\bar c).
  \end{aligned}
  \label{eq:supp-two-aggregation-rules}
\end{equation}

\subsection{Second-Moment Representation}

An individual update can be written as
\begin{equation}
  M(c)
  =sU(c)V(c)^\top
  =s\sum_{q=1}^{r}A_qcc^\top B_q^\top.
  \label{eq:supp-proposal-column-form}
\end{equation}
Define the linear operator
$\mathcal L:\mathbb R^{p\times p}\rightarrow\mathbb R^{o\times d}$ by
\begin{equation}
  \mathcal L(G)=s\sum_{q=1}^{r}A_qGB_q^\top.
  \label{eq:supp-linear-operator}
\end{equation}
For $G_{\mathcal S}=S^{-1}\sum_i c_ic_i^\top$, linearity gives
\begin{align}
  \bar M^{(\mathrm{D\to A})}
  &=\frac{1}{S}\sum_{i=1}^{S}M(c_i) \\
  &=s\sum_{q=1}^{r}A_q
    \left(\frac{1}{S}\sum_i c_ic_i^\top\right)B_q^\top \\
  &=\mathcal L(G_{\mathcal S}),
\end{align}
which proves the second-moment representation for decode--aggregate.

\subsection{Mean--Covariance Decomposition}

Let
\begin{equation}
  \Sigma_{\mathcal S}
  =\frac{1}{S}\sum_i(c_i-\bar c)(c_i-\bar c)^\top.
\end{equation}
Expanding the centered products yields
$G_{\mathcal S}=\bar c\bar c^\top+\Sigma_{\mathcal S}$. Therefore,
\begin{align}
  \bar M^{(\mathrm{D\to A})}
  &=\mathcal L(G_{\mathcal S}) \\
  &=\mathcal L(\bar c\bar c^\top)
    +\mathcal L(\Sigma_{\mathcal S}) \\
  &=\bar M^{(\mathrm{A\to D})}
    +\mathcal L(\Sigma_{\mathcal S}),
  \label{eq:supp-mean-covariance-decomposition}
\end{align}
where the final equality follows from
Eq.~\eqref{eq:supp-proposal-column-form}. This identity compares the two
operations under the same learned maps.

% An equivalent factor-space expression makes the support-variation term
% explicit. Let $U_i=U(c_i)$, $V_i=V(c_i)$,
% $\bar U=S^{-1}\sum_iU_i$, and $\bar V=S^{-1}\sum_iV_i$. Then
% \begin{align}
%   \bar M^{(\mathrm{D\to A})}
%   -\bar M^{(\mathrm{A\to D})}
%   &=\frac{s}{S}\sum_i U_iV_i^\top-s\bar U\bar V^\top \\
%   &=\frac{s}{S}\sum_i
%     (U_i-\bar U)(V_i-\bar V)^\top.
%   \label{eq:supp-factor-cross-covariance}
% \end{align}
% Thus, the additional decode--aggregate term can also be viewed as a learned
% cross-covariance between the generated factor matrices.

\subsection{Rank Bounds and Attainability}

Aggregate--decode produces a single width-$r$ factorization, so
\begin{equation}
  \operatorname{rank}
  \bigl(\bar M^{(\mathrm{A\to D})}\bigr)
  \le r.
  \label{eq:supp-aggregate-decode-rank}
\end{equation}
For decode--aggregate, let $t=\operatorname{rank}(G_{\mathcal S})$. From
Eq.~\eqref{eq:supp-linear-operator},
\begin{align}
  \operatorname{rank}
  \bigl(\bar M^{(\mathrm{D\to A})}\bigr)
  &\le \sum_{q=1}^{r}
    \operatorname{rank}(A_qG_{\mathcal S}B_q^\top) \\
  &\le rt.
\end{align}
Since $G_{\mathcal S}$ is a sum of $S$ rank-one matrices in
$\mathbb R^{p\times p}$, $t\le\min(S,p)$. Combining these facts with the
matrix dimensions yields
\begin{equation}
  \begin{aligned}
    \operatorname{rank}
    \bigl(\bar M^{(\mathrm{D\to A})}\bigr)
    &\le \min\{o,d,r\operatorname{rank}(G_{\mathcal S})\} \\
    &\le \min\{o,d,r\min(S,p)\}.
  \end{aligned}
  \label{eq:supp-decode-aggregate-rank}
\end{equation}
If the two factor branches use bottleneck widths $b_U$ and $b_V$, respectively,
the same argument strengthens the $rt$ term to
$r\min(t,b_U,b_V)$.

The decode--aggregate $rt$ bound is tight when dimensions and decoder capacity
permit. Let $G_{\mathcal S}$ have rank $t$. For each $q$, choose $A_q$ and
$B_q$ to map the $t$-dimensional range of $G_{\mathcal S}$ into disjoint
$t$-dimensional coordinate blocks of the output and input spaces. If
$o,d\ge rt$, each term $A_qG_{\mathcal S}B_q^\top$ then has rank $t$ on a
separate block, so their sum has rank $rt$. In particular, attaining $Sr$
requires $\operatorname{rank}(G_{\mathcal S})=S$, which requires $p\ge S$ and
linearly independent support codes, in addition to sufficient slot dimensions
and, for the bottleneck implementation, $b_U,b_V\ge S$. This construction
establishes architectural capacity only; it does not imply that a trained
model uses the maximum available rank or that greater rank improves
prediction.

% \subsection{Factor-Space Interpretation}

% Linearity of the factor branches gives
% \begin{equation}
%   U(\bar c)=\frac{1}{S}\sum_iU(c_i)=\bar U,
%   \qquad
%   V(\bar c)=\frac{1}{S}\sum_iV(c_i)=\bar V.
% \end{equation}
% Consequently,
% \begin{align}
%   \bar M^{(\mathrm{A\to D})}
%   &=s\bar U\bar V^\top
%     =\frac{s}{S^2}\sum_{i=1}^{S}\sum_{j=1}^{S}U_iV_j^\top, \\
%   \bar M^{(\mathrm{D\to A})}
%   &=\frac{s}{S}\sum_{i=1}^{S}U_iV_i^\top.
% \end{align}
% The first expansion contains $S(S-1)$ cross-example products with $i\ne j$
% and reweights the paired terms. These terms arise algebraically from
% multiplying the empirical factor means; the implementation does not construct
% them explicitly. This is a representational distinction, not a claim that
% cross-example products are necessarily detrimental.

\subsection{Permutation Invariance and Retained Set Statistics}

Both $G_{\mathcal S}=S^{-1}\sum_i c_ic_i^\top$ and
$\bar c=S^{-1}\sum_i c_i$ are unchanged by any permutation of the support
examples. Therefore, both synthesis rules are permutation invariant. Under
fixed learned maps,
\begin{align}
  G_{\mathcal S}=G_{\mathcal S'}
  &\quad\Longrightarrow\quad
  \bar M^{(\mathrm{D\to A})}(\mathcal S)
  =\bar M^{(\mathrm{D\to A})}(\mathcal S'), \\
  \bar c_{\mathcal S}=\bar c_{\mathcal S'}
  &\quad\Longrightarrow\quad
  \bar M^{(\mathrm{A\to D})}(\mathcal S)
  =\bar M^{(\mathrm{A\to D})}(\mathcal S').
\end{align}
Thus decode--aggregate cannot distinguish support sets with the same learned
second moment, while aggregate--decode cannot distinguish sets with the same
learned mean code. These statements concern aggregation in learned code space:
the support conditioner that maps each frozen image embedding to $c_i$ remains
nonlinear, so the mean code can encode richer statistics than the mean input
embedding.

%% file: sections/supplementary_experiments.tex
\section{Experimental Protocol and Implementation}
\label{sec:supp-protocol}

This supplement gives protocol details and expanded results for
Proposal2Adapter. Unless stated otherwise, uncertainties are sample standard
deviations over independently trained seeds.

\subsection{WILDS episodic protocol}

Each source episode contains $S=16$ support and $Q=16$ query images from one
domain, with disjoint image membership. The conditioner receives neither
support labels nor domain, sequence, order, or query information. Query labels
provide cross-entropy for classification and scaled MSE for PovertyMap. At test time, one adapter is
generated from unlabeled target-domain support and applied to that domain's
queries. Checkpoints are selected on OOD validation, frozen before OOD-test
access, and evaluated under immutable test locks. 

For the correct--wrong control, each query cohort is evaluated with its own
domain's adapter and with an adapter generated from an equal-size support set
from a different domain.  Thus query identity and support
count are fixed, and only the conditioning source changes.

\subsection{Source-only baselines and calibration}
\label{sec:supp-source-baselines}

The ViT-B/16 source-only controls in the main-paper comparison use the same frozen
task heads, source eligibility, preprocessing, and official metrics as the
corresponding Proposal2Adapter campaigns, but consume no target support.
ERM updates the complete visual encoder, including its projection, while
keeping the task head frozen. The LoRA control freezes the base encoder and
trains one rank-16, $\alpha=32$ adapter over the fused
attention qkv and output projection in every visual block; it has neither a
conditioner nor a hypernetwork.

Both controls use five seeds, a logical batch size of 256, and 20
image-equivalent source passes. Source batches are domain-balanced: every
64-image sampling group draws four distinct domains uniformly and 16 labeled
images per domain. Training uses AdamW, 500 warm-up updates, cosine decay,
gradient clipping at 1.0, and BF16. Weight decay is 0.1 for ERM and 0.01 for
LoRA. Seed 0 screens
$\{10^{-6},3\!\times\!10^{-6},10^{-5},3\!\times\!10^{-5}\}$ for ERM and
$\{3\!\times\!10^{-5},10^{-4},3\!\times\!10^{-4},10^{-3}\}$ for LoRA using
only the official OOD-validation primary metric. The selected learning rate is
then fixed for all five seeds, and each seed's checkpoint is selected on that
same validation metric before locked OOD-test evaluation. These main-paper
controls use no extra or oracle data. Table~\ref{tab:supp-source-baseline-lrs}
records their selected learning rates. Additional iWildCam backbone, rank,
and extra-camera controls are reported in
Section~\ref{sec:supp-backbone-source-controls}.

\begin{table}[t]
  \centering
  \caption{Validation-selected learning rates for the source-only controls.}
  \label{tab:supp-source-baseline-lrs}
  \setlength{\tabcolsep}{5pt}
  \small
  \begin{tabular}{lcc}
    \toprule
    Dataset & ERM & LoRA ($r=16$) \\
    \midrule
    iWildCam & $3\!\times\!10^{-6}$ & $3\!\times\!10^{-4}$ \\
    FMoW & $10^{-5}$ & $10^{-4}$ \\
    PovertyMap & $3\!\times\!10^{-6}$ & $10^{-4}$ \\
    \bottomrule
  \end{tabular}
\end{table}

\subsection{Support representation and adapter generation}

Let $f$ denote the frozen CLIP image encoder and $x_i$ a support image. We use
the final normalized embedding
\begin{equation}
  z_i=f(x_i)
\end{equation}
and form the per-example support code
\begin{equation}
  c_i=F_c(P_z(z_i)).
\end{equation}
$P_z$ has width 128. The shared code network $F_c$ uses a 512-wide hidden layer
for ViT-B/16 and a 768-wide layer for ViT-L/14, and outputs a 128-D code. Each
slot factor decoder contains two bias-free bottleneck branches of width 32,
which independently generate the $U$ and $V$ factors. In the canonical
decode--aggregate configuration, each support code is decoded separately and
the completed dense updates are averaged. In aggregate--decode, the support
codes are averaged before a single factor-decoder evaluation per slot. The two
variants have identical trainable architectures and parameter counts but are
initialized and trained independently.

ViT-B/16 uses the final normalized 512-D CLIP embedding and 24 attention-LoRA
slots. ViT-L/14 uses its normalized 768-D embedding and 48 slots. Both use
rank 16 and LoRA scale $\alpha=32$. The CLIP backbone and task head---a prompt
classifier or the PovertyMap regressor---are frozen; the conditioner and decoder bank are initialized from
scratch in every definitive run.

\subsection{Input preprocessing}

Images are converted to RGB, resized bicubically so that the shortest edge is
224 pixels, center-cropped to $224\times224$, and normalized with the pinned
CLIP statistics. The final L2-normalized frozen CLIP embedding is the support
descriptor.

\subsection{Optimization}

Table~\ref{tab:supp-training-recipes} summarizes the training recipes for all
reported Proposal2Adapter WILDS campaigns.
All Proposal2Adapter WILDS campaigns use AdamW, weight decay 0.01, cosine decay, gradient clipping at
1.0, and BF16 training. The coarse/fine comparisons in
Section~\ref{sec:supp-domain-granularity} are separate matched 20K-step studies
and were not repeated for the headline two-order campaign.

\begin{table*}[t]
  \centering
  \caption{Training recipes for the reported campaigns. ``2+2'' denotes two
  ordinary and two extra-oracle episodes per update.}
  \label{tab:supp-training-recipes}
  \setlength{\tabcolsep}{4.3pt}
  \small
  \begin{tabular}{llrrrrrr}
    \toprule
    Campaign & Backbone & Steps & Episodes/update & LR & Warm-up & Val. interval & Seeds \\
    \midrule
    iWildCam standard and camera ladders & ViT-B/16 & 20K & 4 & $4\!\times\!10^{-4}$ & 500 & 250 & 5 \\
    iWildCam extra-oracle & ViT-B/16 & 20K & 2+2 & $4\!\times\!10^{-4}$ & 500 & 250 & 5 \\
    iWildCam aggregation order & ViT-B/16 & 20K & 4 or 2+2 & $4\!\times\!10^{-4}$ & 500 & 250 & 5 \\
    iWildCam standard/oracle & ViT-L/14 & 20K & 4 or 2+2 & $10^{-4}$ & 500 & 250 & 5 \\
    FMoW headline & ViT-B/16 & 20K & 4 & $4\!\times\!10^{-4}$ & 500 & 250 & 5 \\
    PovertyMap headline & ViT-B/16 & 20K & 4 & $4\!\times\!10^{-4}$ & 500 & 250 & 5 \\
    \bottomrule
  \end{tabular}
\end{table*}

\subsection{iWildCam aggregation-order ablation}
\label{sec:supp-aggregation-order}

Table~\ref{tab:supp-aggregation-order-test} gives the matched-seed OOD-test
results underlying the aggregation-order table in the main paper.
The extra-camera setting retains the original labeled pool and adds all 1,264
eligible cameras using labels normally hidden by the benchmark.
Aggregate--decode minus decode--aggregate macro-F1 is
$+1.40$ points with a descriptive paired 95\% interval of
$[-0.70,3.49]$ under ordinary supervision, and $-0.83$ points with interval
$[-3.13,1.47]$ with the 1,264 extra labeled cameras. The corresponding
accuracy and conditioning-gap contrasts also have intervals spanning zero.
These intervals describe five matched training seeds. 

\begin{table*}[t]
  \centering
  \caption{Per-seed locked iWildCam OOD-test results for aggregation order.
  A--D$-$D--A is the paired macro-F1 difference. Gap denotes the
  correct-minus-wrong support macro-F1 difference. Scores are percentage
  points.}
  \label{tab:supp-aggregation-order-test}
  \setlength{\tabcolsep}{2.7pt}
  \scriptsize
  \begin{tabular}{llrrrrrrr}
    \toprule
    Source & Seed & D--A F1 & A--D F1 & A--D$-$D--A & D--A Acc. & A--D Acc. & D--A gap & A--D gap \\
    \midrule
    Ordinary & 0 & 39.49 & 39.43 & $-0.06$ & 72.48 & 75.78 & 22.38 & 22.95 \\
    Ordinary & 1 & 37.13 & 37.95 & $+0.82$ & 75.60 & 76.00 & 23.34 & 22.71 \\
    Ordinary & 2 & 35.80 & 39.01 & $+3.21$ & 74.28 & 76.36 & 18.89 & 24.02 \\
    Ordinary & 3 & 39.64 & 39.47 & $-0.17$ & 76.59 & 73.57 & 24.18 & 23.87 \\
    Ordinary & 4 & 37.07 & 40.25 & $+3.18$ & 75.61 & 75.09 & 20.24 & 24.15 \\
    \midrule
    +1,264 extra & 0 & 43.55 & 43.25 & $-0.30$ & 75.95 & 77.71 & 27.13 & 28.22 \\
    +1,264 extra & 1 & 44.54 & 43.75 & $-0.79$ & 78.46 & 74.65 & 26.93 & 27.55 \\
    +1,264 extra & 2 & 45.88 & 43.67 & $-2.21$ & 76.50 & 76.33 & 30.15 & 30.55 \\
    +1,264 extra & 3 & 44.57 & 41.77 & $-2.80$ & 70.04 & 73.96 & 27.30 & 25.95 \\
    +1,264 extra & 4 & 43.40 & 45.34 & $+1.94$ & 74.64 & 76.15 & 27.11 & 28.97 \\
    \bottomrule
  \end{tabular}
\end{table*}

% \paragraph{Update-geometry computation.}
% For the geometry summaries in main-paper Table~2, we materialize each
% test-support update in FP32 and compute its full SVD in FP64. Numerical rank
% counts singular values above
% $\max(d_{\mathrm{out}},d_{\mathrm{in}})\epsilon_{\mathrm{FP32}}\sigma_{\max}$,
% and entropy rank is $\exp[-\sum_jp_j\log p_j]$ for
% $p_j=\sigma_j/\sum_k\sigma_k$. We average slots within each camera and cameras
% within each seed before reporting the five-seed mean and sample standard
% deviation; cameras and slots are descriptive observations, not independent
% replicates.

% Both orders improve in every matched seed when ordinary supervision is
% augmented with the 1,264 extra labeled cameras. The mean gains are $+6.56$
% points for D--A and $+4.33$ for A--D. Their paired difference in gains is $2.23$ points with a
% descriptive 95\% interval of $[-0.23,4.69]$; it is suggestive rather than
% conclusive evidence that D--A benefits more from richer source supervision.

\section{Source-Domain Scale and Specialization: Expanded Results}
\label{sec:supp-source-scale}

\subsection{iWildCam camera ladders and DA-MergeLoRA source-scale control}
\label{sec:supp-source-scale-iwildcam}

For the ordinary labeled pool, a camera is episode-eligible if it contains at least \(S+Q=32\) training images, permitting sample-disjoint \(S=Q=16\) sampling without replacement. This retains 212 of 243 training cameras and 129,394 of 129,809 images; the remaining 31 cameras contain only 415 images in total. 

The Random and Class-stratified ladders contain nested subsets of
53/106/159/212 cameras drawn from 212 eligible original cameras. The Random
ladder changes both domain count and class coverage. Class-stratified selection
retains all 174 source-present classes at every budget while selecting whole
cameras and approximating the full-pool sampler-weighted class distribution.
It does not balance the training loss. Selected image counts average
45,775/75,097/102,476/129,394. Fifty-seven
classes occur in only one camera, so class completeness constrains the small
subsets strongly.

The extra-labeled oracle ladder keeps all 212 original cameras and adds nested
subsets of 316/632/948/1,264 cameras. The eligible extra pool contains 693,333
images in 98,497 sequences and represents 128 of the 182 task labels. Each
camera must contain at least 32 images with $y\in\{0,\ldots,181\}$; images with
$y=-1$ or $y=99999$ are excluded. This permits sample-disjoint $S=Q=16$
sampling without replacement and does not impose sequence-disjointness. Each
training update contains two ordinary and two oracle supervised episodes.
Because these labels are normally hidden, the ladder is only a diagnostic of
additional source supervision. Camera count is also confounded with image
volume, sequence count, class frequencies, and source-mixture changes.

\paragraph{Both aggregation orders.}
Table~\ref{tab:supp-both-order-camera-ladders} adds the complete
aggregate--decode camera ladders to the decode--aggregate study reported in
the main paper.  The two orders use the same source-camera subsets at every
seed and budget.  The correct--wrong gap evaluates the same query cohort with
the camera's own support and with cardinality-matched support from a different
camera; it therefore measures functional dependence on the conditioning
camera rather than classification performance alone.

\begin{table*}[t]
  \centering
  \caption{Locked iWildCam OOD-test camera ladders for decode--aggregate
  (D--A) and aggregate--decode (A--D). Entries are five-seed mean (sample SD)
  on the 0--100 scale. Gap is the correct-minus-wrong-support macro-F1
  difference on its separate cardinality-matched 44-camera query cohort.
  Random and Class-stratified select from 212 eligible ordinary
  cameras. The Extra ladder retains those 212 and adds the indicated number
  of eligible extra cameras. The ordinary 100\% and Extra 0\% rows reuse the
  same endpoint within each order.}
  \label{tab:supp-both-order-camera-ladders}
  \setlength{\tabcolsep}{3.2pt}
  \scriptsize
  \begin{tabular}{llrcccc}
    \toprule
    Ladder & Budget & Source cameras & D--A F1 & A--D F1 & D--A gap & A--D gap \\
    \midrule
    Random & 25\%  & 53  & 26.47 (0.98) & 26.36 (2.50) & 11.29 (1.39) & 10.15 (2.47) \\
    Random & 50\%  & 106 & 30.93 (1.70) & 32.11 (1.56) & 16.27 (1.62) & 17.00 (1.72) \\
    Random & 75\%  & 159 & 35.11 (2.17) & 35.72 (2.41) & 20.05 (2.29) & 19.88 (1.73) \\
    Random & 100\% & 212 & 37.82 (1.67) & 39.22 (0.84) & 21.81 (2.19) & 23.54 (0.66) \\
    \midrule
    Class-stratified & 25\%  & 53  & 30.88 (1.30) & 31.45 (1.54) & 12.66 (1.86) & 13.22 (1.91) \\
    Class-stratified & 50\%  & 106 & 33.45 (0.61) & 33.21 (1.16) & 17.73 (1.11) & 16.55 (2.26) \\
    Class-stratified & 75\%  & 159 & 35.54 (1.61) & 35.96 (2.96) & 20.15 (2.11) & 20.10 (3.32) \\
    Class-stratified & 100\% & 212 & 37.82 (1.67) & 39.22 (0.84) & 21.81 (2.19) & 23.54 (0.66) \\
    \midrule
    Extra & 0\%   & $212+0$     & 37.82 (1.67) & 39.22 (0.84) & 21.81 (2.19) & 23.54 (0.66) \\
    Extra & 25\%  & $212+316$   & 40.26 (1.72) & 41.42 (1.02) & 25.04 (1.81) & 25.87 (0.95) \\
    Extra & 50\%  & $212+632$   & 42.85 (2.33) & 41.39 (2.10) & 27.02 (2.63) & 25.39 (1.81) \\
    Extra & 75\%  & $212+948$   & 42.45 (0.68) & 43.13 (0.70) & 26.44 (1.42) & 26.43 (0.89) \\
    Extra & 100\% & $212+1{,}264$ & 44.39 (0.99) & 43.56 (1.27) & 27.72 (1.36) & 28.25 (1.70) \\
    \bottomrule
  \end{tabular}
\end{table*}

Aggregate--decode independently recovers the source-scale trend. From 25\% to
100\% ordinary cameras, macro-F1 rises by 12.86 points on the Random ladder
and 7.77 points on the Class-stratified ladder; the corresponding
decode--aggregate gains are 11.35 and 6.95 points. Adding all 1,264 extra
cameras raises aggregate--decode by 4.33 points and decode--aggregate by 6.56
points, with positive endpoint changes in all five seeds for both orders.
Every ladder/order also has a wider endpoint conditioning gap. Intermediate
Extra-ladder means are not strictly monotone, so we interpret the full-pool
contrast rather than a smooth dose response.

The extra-camera gain is 2.23 points larger for decode--aggregate than for
aggregate--decode, and all five paired seed differences have that direction;
the descriptive paired 95\% interval is $[-0.23,4.69]$. The ordinary Random
and Class-stratified endpoint gains instead slightly favor aggregate--decode.
The present evidence therefore weakly favors decode--aggregate for this
particular extra-camera expansion, but does not identify one aggregation order
as uniformly more source-scalable.

\paragraph{DA-MergeLoRA source-scale control.}
We also evaluate a pinned, released-style implementation of
DA-MergeLoRA~\cite{reid2026damerge} with $K\in\{10,20\}$ source clusters.
Each condition uses OpenAI CLIP ViT-B/16, the frozen
182-way prompt head, rank-16 ($\alpha=32$) attention adapters, and five matched
training seeds. The full ordinary pool contains 243 cameras; the half-pool
contains 122 seed-specific cameras; and the extra condition jointly clusters
all 243 ordinary and 1,264 extra cameras. Source adapters are fit for five
image epochs per cluster, followed by 2,000 merger updates. Fixed final states
are evaluated on the locked target protocol without test-based selection.
Table~\ref{tab:supp-damerge-source-scale} reports the per-seed locked-test
scores and conditioning gaps.

\begin{table*}[t]
  \centering
  \caption{DA-MergeLoRA source-scale control on the locked iWildCam OOD test.
  Seed columns and Mean (SD) report full-domain macro-F1 on the 0--100 scale;
  Gap is the five-seed correct-minus-wrong-support macro-F1 mean (SD).}
  \label{tab:supp-damerge-source-scale}
  \setlength{\tabcolsep}{3.4pt}
  \scriptsize
  \begin{tabular}{clrrrrrrcc}
    \toprule
    $K$ & Source & Cameras & Seed 0 & Seed 1 & Seed 2 & Seed 3 & Seed 4 & Mean (SD) & Gap \\
    \midrule
    10 & Ordinary 50\% & 122   & 27.58 & 28.80 & 30.59 & 29.83 & 29.07 & 29.17 (1.13) & 13.99 (1.06) \\
    10 & Ordinary 100\% & 243  & 34.41 & 33.02 & 33.47 & 32.94 & 32.42 & 33.25 (0.75) & 17.86 (1.48) \\
    10 & +1,264 extra & 1,507   & 34.35 & 36.06 & 36.15 & 36.27 & 36.62 & 35.89 (0.89) & 19.91 (1.07) \\
    \midrule
    20 & Ordinary 50\% & 122   & 29.39 & 28.72 & 28.46 & 27.34 & 27.58 & 28.30 (0.84) & 13.56 (1.71) \\
    20 & Ordinary 100\% & 243  & 31.28 & 32.94 & 33.34 & 32.20 & 33.08 & 32.57 (0.84) & 18.57 (0.72) \\
    20 & +1,264 extra & 1,507   & 33.79 & 32.22 & 32.80 & 33.44 & 32.45 & 32.94 (0.66) & 18.12 (1.42) \\
    \bottomrule
  \end{tabular}
\end{table*}

At $K=10$, moving from half to all ordinary cameras improves macro-F1 by
$4.08\pm1.62$ points (five of five positive seeds), while adding the extra
cameras to the full ordinary pool improves it by $2.64\pm1.61$ points (four of
five positive; descriptive 95\% interval $[0.64,4.64]$). At $K=20$, the
corresponding gains are $4.27\pm1.41$ points (five of five) and
$0.38\pm1.44$ points (two of five; $[-1.42,2.17]$). The full-ordinary $K=10$
mean of 33.25 is close to the saved published anchor of 33.62, providing an
implementation sanity check. The sensitivity to $K$ is itself informative:
the extra-camera benefit is not stable to increasing the number of source
clusters.

The larger observed Proposal2Adapter gains in
Table~\ref{tab:supp-key-source-richness} support the source-richness
hypothesis, but they are not a controlled method-by-source interaction.
Proposal2Adapter samples 212 episode-eligible ordinary cameras uniformly and
uses a fixed 2+2 ordinary/extra mixture at the added-camera endpoint.
DA-MergeLoRA uses all 243 ordinary cameras, recomputes the joint clustering,
refits every source adapter, and changes both image exposure and optimization
work. The comparison therefore establishes a stronger observed scaling pattern
under the respective protocols, not a causal attribution to architecture
alone.

\subsection{Backbone scale, aggregation order, and source-only controls}
\label{sec:supp-backbone-source-controls}

Table~\ref{tab:supp-backbone-source-controls} compares both aggregation orders
with ERM and static LoRA at ranks 16, 24, and 32. All methods use the same
frozen prompt head and source eligibility; the extra-camera setting retains
the 212 ordinary cameras and adds the 1,264 eligible cameras defined above.
Both source-only and Proposal2Adapter training use equal ordinary/extra loss
weight in this setting. Source baselines follow
Section~\ref{sec:supp-source-baselines}, with an independent validation-only
LR screen for each backbone, rank, and source setting and $\alpha=2r$ for
LoRA. Every iWildCam LoRA family selects $3\!\times\!10^{-4}$; ERM selects
$3\!\times\!10^{-6}$ except for ViT-L/14 with extra cameras, which selects
$10^{-5}$. The baseline budget is 20 image-equivalent passes (10,109 updates
for ordinary sources and 64,276 with extra cameras), whereas Proposal2Adapter
uses 20K episodic updates in both settings; these are not compute-matched
comparisons. ViT-L/14 aggregate--decode inherits the decode--aggregate recipe
without additional tuning.

\begin{table*}[t]
  \centering
  \caption{iWildCam source-only controls and both Proposal2Adapter aggregation
  orders across backbone and source settings. Entries are full-domain OOD-test
  macro-F1 on the 0--100 scale, reported as five-seed mean (sample SD).
  ``+1,264 extra'' adds normally hidden labels for the eligible extra cameras
  and is an oracle-supervision diagnostic. D--A denotes decode--aggregate and
  A--D aggregate--decode; both use factor rank $r=16$. All task heads are
  frozen. The training budgets differ as described in the text.}
  \label{tab:supp-backbone-source-controls}
  \setlength{\tabcolsep}{3.0pt}
  \small
  \begin{tabular}{@{}llcccccc@{}}
    \toprule
    Backbone & Source & ERM & \multicolumn{3}{c}{Static LoRA} & \multicolumn{2}{c}{Proposal2Adapter} \\
    \cmidrule(lr){4-6}\cmidrule(l){7-8}
     & & & $r=16$ & $r=24$ & $r=32$ & D--A & A--D \\
    \midrule
    ViT-B/16 & Ordinary & 35.21 (1.17) & 31.83 (1.04) & 33.40 (0.60) & 33.73 (1.18) & 37.82 (1.67) & 39.22 (0.84) \\
     & +1,264 extra & 40.37 (0.37) & 37.45 (1.26) & 39.67 (0.44) & 37.91 (1.27) & 44.39 (0.99) & 43.56 (1.27) \\
    \midrule
    ViT-L/14 & Ordinary & 41.06 (1.84) & 41.17 (1.20) & 40.07 (0.90) & 41.06 (0.36) & 44.85 (1.37) & 44.44 (0.33) \\
     & +1,264 extra & 48.52 (1.28) & 46.27 (0.62) & 45.66 (2.25) & 45.83 (1.88) & 49.25 (1.10) & 49.03 (1.55) \\
    \bottomrule
  \end{tabular}
\end{table*}

Both aggregation orders outperform the ordinary-source ERM and static-LoRA
controls with either backbone; increasing static LoRA rank does not close the
gap. With ViT-L/14, extra cameras improve D--A and A--D by 4.40 and 4.59
macro-F1 points, respectively, with no clear order advantage. Full visual ERM
becomes competitive under expanded supervision, reaching 48.52 compared with
49.25 and 49.03 for the two orders, while static LoRA remains lower across
the tested ranks. Thus, the source-richness benefit extends to both orders,
while their margin over full fine-tuning depends on backbone and supervision.

\subsection{Domain granularity}
\label{sec:supp-domain-granularity}

FMoW compares coarse year domains with fine region--year domains at a matched
20K-step budget. PovertyMap compares 13 country domains with 26
country--urban domains. The PovertyMap validation/test reversal and the lack of
correct--wrong controls make its result descriptive rather than evidence of
increased specialization. Table~\ref{tab:supp-domain-granularity} reports the
matched coarse-versus-fine comparisons.

\begin{table*}[t]
  \centering
  \caption{Effect of fine versus coarse source-domain definitions. Entries are
  official metrics; $\Delta$ is fine minus coarse, paired by seed. FMoW reports
  worst-region accuracy and PovertyMap worst-group Pearson $r$.}
  \label{tab:supp-domain-granularity}
  \setlength{\tabcolsep}{2.0pt}
  \scriptsize
  \begin{tabular}{llcccc}
    \toprule
    Dataset/split & Method & Coarse & Fine & $\Delta$ & Correct--wrong (coarse $\to$ fine) \\
    \midrule
    FMoW val & Proposal2Adapter (D--A) & $.530511\pm.003295$ & $.554172\pm.006471$ & $+.023661\pm.003295$ & -- \\
    FMoW test & Proposal2Adapter (D--A) & $.380897\pm.016589$ & $.430775\pm.022724$ & $+.049878\pm.039311$ & $-.001028\to+.065047$ \\
    \midrule
    PovertyMap val & Proposal2Adapter (D--A) & -- & -- & $-.029492\pm.016618$ & unavailable \\
    PovertyMap test & Proposal2Adapter (D--A) & $.533545\pm.055267$ & $.545423\pm.011420$ & $+.011878\pm.047004$ & unavailable \\
    \bottomrule
  \end{tabular}
\end{table*}

The FMoW results show that coherent, support-identifiable domains matter more
than domain count alone: fine domains improve both worst-region accuracy and
the functional conditioning effect. PovertyMap remains inconclusive because
finer domains reduce validation performance while improving the locked-test
point estimate. This instability may reflect insufficient samples per domain,
heterogeneous country--urban groups, or selection noise; the current
experiment does not distinguish these explanations.

\section{Target-Support-Size Details}
\label{sec:supp-support-size}

This post-hoc analysis reuses the independently trained, validation-selected
16-shot iWildCam checkpoints for both aggregation orders; no model is retrained
or reselected. At inference, support caps are reduced through nested per-camera
prefixes. The primary query cohort remains fixed at 42,791 images from 48
cameras. The matched wrong-support cohort varies slightly with the cap, but
correct and wrong support use identical queries within each cell. Caps
$16/8/4/2/1$ realize $739/378/191/96/48$ support images across the 48 target
cameras.

\begin{table*}[t]
  \centering
  \caption{Support-size sensitivity and realized update geometry for both
  Proposal2Adapter aggregation orders on the iWildCam OOD test. Macro-F1 and
  $\Delta_{\mathrm{cond}}$ are on the 0--100 scale; the latter is the matched
  correct-minus-wrong-support macro-F1 gap. Entries are mean (sample standard
  deviation) over five matched seeds. $E_{>16}$ is squared spectral energy
  beyond rank 16; zeros are exact up to floating-point SVD residuals.}
  \label{tab:supp-support-size}
  \setlength{\tabcolsep}{4.0pt}
  \small
  \begin{tabular}{rccccc}
    \toprule
    Cap & Order & Macro-F1 & $\Delta_{\mathrm{cond}}$ & Entropy rank & $E_{>16}$ (\%) \\
    \midrule
    16 & D--A & 37.82 (1.67) & 21.81 (2.19) & 29.36 (1.39) & 3.05 (0.52) \\
    16 & A--D & 39.22 (0.84) & 23.54 (0.66) & 15.06 (0.01) & 0 \\
     8 & D--A & 37.98 (1.77) & 22.47 (2.89) & 27.59 (0.73) & 3.08 (0.32) \\
     8 & A--D & 38.81 (0.70) & 23.46 (1.19) & 15.06 (0.02) & 0 \\
     4 & D--A & 36.85 (1.59) & 20.29 (3.20) & 24.67 (0.58) & 3.00 (0.38) \\
     4 & A--D & 38.16 (0.77) & 21.36 (1.06) & 15.07 (0.02) & 0 \\
     2 & D--A & 35.57 (1.76) & 18.42 (3.24) & 20.13 (0.37) & 2.18 (0.35) \\
     2 & A--D & 37.25 (0.95) & 20.07 (0.50) & 15.08 (0.02) & 0 \\
     1 & D--A & 32.85 (1.14) & 17.18 (1.31) & 15.17 (0.04) & 0 \\
     1 & A--D & 34.00 (1.02) & 18.23 (1.85) & 15.11 (0.01) & 0 \\
    \bottomrule
  \end{tabular}
\end{table*}

Table~\ref{tab:supp-support-size} combines prediction, conditioning, and
update-geometry results at every support cap. Aggregate--decode has the higher
mean macro-F1 at every cap, with A--D minus D--A differences of 1.40, 0.83,
1.31, 1.67, and 1.15 points from 16 to one shot. The post-hoc, unadjusted
paired 95\% intervals exclude zero only at two
shots, $[0.11,3.23]$, and one shot, $[0.31,1.99]$. All paired
difference-in-change intervals include zero: the orders have essentially the
same relative macro-F1 sensitivity, retaining 97.29\% (A--D) and 97.42\%
(D--A) at four shots and 86.70\% and 86.86\% at one shot. Correct support
outperforms matched wrong-camera support in every order--cap--seed cell.

Decode--aggregate's entropy rank contracts from 29.36 to 15.17 as support
decreases from 16 to one image, while aggregate--decode remains near 15.1.
At one shot the two synthesis formulas have the same single-example form and
both updates have numerical rank 16, yet the independently trained A--D models
retain a 1.15-point mean advantage. Thus the sweep verifies the predicted
support-dependent rank behavior without implying that realized rank alone
causes the performance difference. The DA-MergeLoRA curve in the main paper is
a cross-paper comparison: it uses three training seeds, reports no cap-2
result, and follows its own training procedure.

%% file: sections/supplementary_libero.tex
\section{Adapting to New Tasks in LIBERO}
\label{sec:supp-libero-protocol}

\subsection{Target-support size}

We evaluate whether additional target demonstrations improve adaptation
at inference. During training, the adapter generator receives 16 images
from one demonstration per episode. We hold the trained checkpoints fixed
and vary inference support from one to three, five, or ten demonstrations,
using 16 images from each.

\paragraph{Protocol.}
We use the Spatial, Object, Goal, and LIBERO-10 (Long) suites of
LIBERO~\cite{liu2023libero}, each with ten tasks. Each fold trains on
three suites and evaluates on the fourth. The image-only adapter generator
builds one adapter per target task from the support demonstrations.
The adapted policy receives the task instruction, live camera views,
and robot state. Full-task success is measured over 50 simulator starts
per task and seed.

\paragraph{Training.}
Query action supervision comes from a separate demonstration of the same
task. We use decode--aggregate synthesis and keep the OpenPI
\texttt{pi05\_base} policy frozen~\cite{black2025pi05}. Generated
rank-16 updates ($\alpha=32$) act on selected language and action
attention projections. Training uses four episodes per update,
20,000 updates, three seeds, and source-validation checkpoint selection.
The conventional LoRA baseline in the main paper fits one fixed adapter
on the same 30 source tasks for 20,000 updates.

\begin{table}[!htb]
  \centering
  \caption{Proposal2Adapter full-task success (\%) by number of
  demonstrations at inference. Entries give the mean (sample SD) across
  three seeds. Bold marks the highest mean in each row.}
  \label{tab:supp-libero-support}
  \setlength{\tabcolsep}{2.1pt}
  \footnotesize
  \begin{tabular}{@{}lrrrr@{}}
    \toprule
    & \multicolumn{4}{c}{Support demonstrations} \\
    \cmidrule(l){2-5}
    Suite & 1 & 3 & 5 & 10 \\
    \midrule
    Spatial & 17.87 (2.32) & 18.20 (2.80) & \textbf{19.73} (2.55) & 19.20 (1.56) \\
    Object  & \textbf{11.60} (6.13) & 10.27 (6.13) & 10.87 (6.54) & 10.20 (5.86) \\
    Goal    & \textbf{8.67} (2.14) & 8.60 (2.65) & 8.40 (1.73) & 8.00 (1.71) \\
    Long    & \textbf{0.33} (0.42) & 0.20 (0.20) & 0.27 (0.23) & 0.13 (0.23) \\
    \bottomrule
  \end{tabular}
\end{table}

Additional support gives a modest improvement on Spatial, from
17.87\% success with one demonstration to 19.73\% with five, followed
by 19.20\% with ten. The other suites show no consistent improvement
over this range (Table~\ref{tab:supp-libero-support}).

\subsection{Expanding source-task supervision}
\label{sec:supp-libero-expansion}

To test the benefit of richer source supervision, we start two runs
for each held-out suite and seed from the same 30-task checkpoint
evaluated in Table~\ref{tab:supp-libero-support}. Both train for a
further 20,000 updates. One continues on the original 30 tasks; the
other adds LIBERO-90, expanding training to 120 tasks across kitchens,
living rooms, and studies. Both use fresh optimizers and share the
original normalization, validation tasks, target demonstrations, and
simulator starts.

\begin{figure}[tbp]
  \centering
  \includegraphics[width=\linewidth]{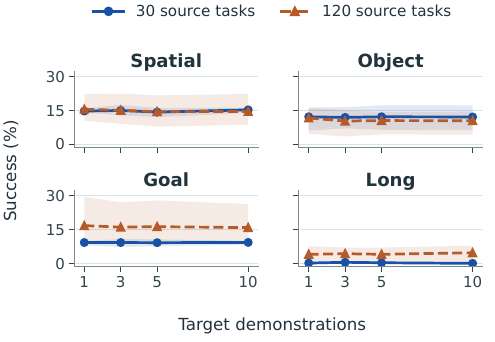}
  \caption{Full-task success after continued training on 30 or 120 source
  tasks. At each demonstration count, the vertical gap compares the two
  training sets using the same target demonstrations. Lines give the mean
  across three seeds and shading spans the individual seed results.}
  \label{fig:supp-libero-expansion}
\end{figure}

\begin{table}[!htb]
  \centering
  \caption{Held-out success (\%) with one target demonstration after
  20,000 additional training updates on 30 or 120 tasks. Each mean
  (sample SD) uses three seeds and 500 rollouts per suite per seed.
  Bold marks the higher mean.
  $\Delta$ gives the paired change in points.}
  \label{tab:supp-libero-expansion}
  \setlength{\tabcolsep}{4pt}
  \small
  \begin{tabular}{@{}lrrr@{}}
    \toprule
    Held-out suite & 30 tasks & 120 tasks & $\Delta$ \\
    \midrule
    Spatial & 14.67 (0.99) & \textbf{15.47} (6.07) & $+0.80$ \\
    Object  & \textbf{12.13} (5.40) & 11.60 (6.13) & $-0.53$ \\
    Goal    &  9.20 (1.04) & \textbf{16.67} (11.04) & $+7.47$ \\
    Long    &  0.13 (0.23) &  \textbf{4.00} (3.82) & $+3.87$ \\
    \bottomrule
  \end{tabular}
\end{table}

With one target demonstration, adding LIBERO-90 raises mean success
from 9.20\% to 16.67\% on Goal and from 0.13\% to 4.00\% on Long,
while changes on Spatial and Object are small
(Table~\ref{tab:supp-libero-expansion}). The expanded-source model
also has higher mean success on Goal and Long at each support size in
Fig.~\ref{fig:supp-libero-expansion}. Thus, additional source tasks
benefit some held-out suites substantially, rather than improving all
suites uniformly.

\subsection{Understanding performance on Long}
\label{sec:supp-libero-partial}

\begin{figure*}[t]
  \centering
  \includegraphics[width=0.98\textwidth]{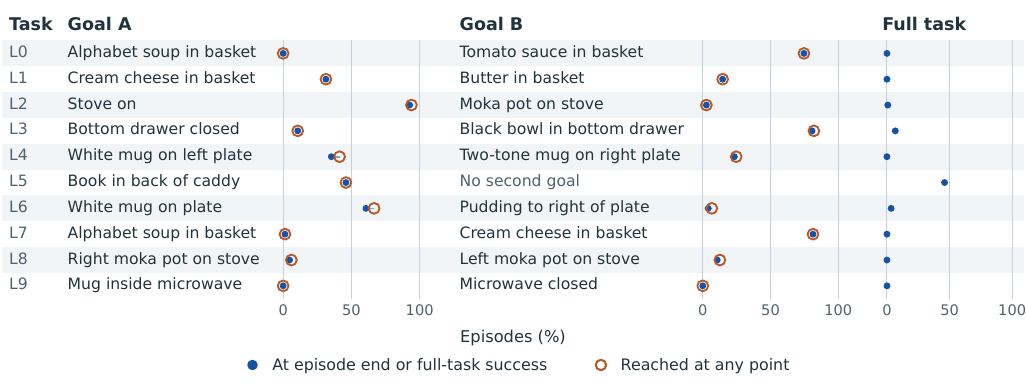}
  \caption{Individual goals and full-task success on LIBERO-10.
  Each row names the goals for one task; L5 has only one.
  On L4, A is the white mug on the left plate and B is the
  yellow-and-white mug on the right plate. Each goal is scored at episode
  end and at any point during the same 150 attempts with the full
  instruction and matching support. The right panel shows full-task
  success. For L8, partial credit counts pot placements; the stove is
  initially on and must remain on for full success.}
  \label{fig:supp-libero-long-components}
\end{figure*}

The book-placement task (L5) has one goal, whereas the other nine Long tasks require
two outcomes. On L4, for example, the white mug belongs on the left
plate and the yellow-and-white mug on the right; full success requires
both. Following WIZARD~\cite{bianchi2026wizard}, we score individual
goals in fresh rollouts. For each held-out suite, including LIBERO-90,
we use a freshly initialized generator trained for 20,000 updates on
the other four suites with the base policy frozen. This gives 120
source tasks for a ten-task holdout and 40 for the LIBERO-90 holdout.
Each target task supplies 50 matching demonstrations of 16 images.
We evaluate 50 starts per task and seed, giving 19,500 episodes across
130 tasks and three seeds.

The policy receives the full instruction on every attempt. We record
which goals are reached at any point and which hold at episode end.
Full-task success requires all simulator conditions. On L8, the stove
starts on: the two pot placements are scored separately, while full
success also requires the stove to remain on.

On L4, the white mug ended on its plate in 53/150 episodes (35.33\%)
and the yellow-and-white mug in 35/150 (23.33\%), but both goals held
together at episode end in 0/150. Across the fresh Long rollouts, 69 of the 85 full
successes came from L5. The other nine tasks supplied 16 full successes
in 1,350 attempts (Fig.~\ref{fig:supp-libero-long-components}).
These rollouts give 5.67\% full-task success on Long; they are separate
from the continued-training experiment in
Table~\ref{tab:supp-libero-expansion}.

\begin{table}[!htb]
  \centering
  \caption{Terminal-state scores (\%), weighting tasks equally.
  Mean goal completion averages the fraction of scored goals satisfied
  per episode; at least one goal counts episodes satisfying any scored
  goal. Full-task success requires all simulator conditions.}
  \label{tab:supp-libero-partial}
  \setlength{\tabcolsep}{3.3pt}
  \footnotesize
  \begin{tabular}{@{}lrrrr@{}}
    \toprule
    Query group & Tasks & \shortstack{Mean goal\\completion} &
    \shortstack{At least\\one goal} & \shortstack{Full-task\\success} \\
    \midrule
    Spatial   & 10 & 17.93 & 17.93 & 17.93 \\
    Object    & 10 & 10.40 & 10.40 & 10.40 \\
    Goal      & 10 & 20.67 & 20.67 & 20.67 \\
    Long      & 10 & 31.03 & 56.40 &  5.67 \\
    LIBERO-90 & 90 & 21.38 & 23.38 & 19.38 \\
    \midrule
    Two goals & 19 & 23.81 & 46.63 &  0.98 \\
    One goal & 111 & 20.47 & 20.47 & 20.47 \\
    All tasks & 130 & 20.96 & 24.29 & 17.62 \\
    \bottomrule
  \end{tabular}
\end{table}

Across all 19 two-goal tasks, at least one goal was satisfied at
episode end in 1,329/2,850 rollouts (46.63\%), while the full instruction
was completed in 28/2,850 (0.98\%; Table~\ref{tab:supp-libero-partial}).
Partial completion therefore reveals capabilities that full-task success
alone does not capture, while jointly satisfying the requested outcomes
remains a substantial limitation.

\begin{center}\begin{minipage}{\columnwidth}
  \centering
  \includegraphics[width=\linewidth]{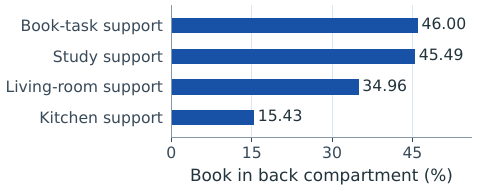}
  \captionof{figure}{Book placement in the back compartment (L5) with matching
  support or individual LIBERO-90 support tasks grouped by room.
  Each support task is evaluated over 150 rollouts; room bars average
  over 17 study, 27 living-room, and 46 kitchen tasks.}
  \label{fig:supp-libero-book}
\end{minipage}\end{center}

\paragraph{Support context for the book task.}
Since L5 accounts for most full-task successes on Long, we examine
which demonstrations support its performance. Using the same trained
generators and support budget, we hold the book-placement query and simulator starts fixed
while varying the support task. On L5, placing a book in the
\emph{back compartment} of a desk caddy, matching support yields
69/150 successes (46.00\%). Support from a LIBERO-90 task that places
the same book in the \emph{left compartment} of the same caddy also
yields 69/150 successes on the back-compartment query. Figure~\ref{fig:supp-libero-book} groups source-support results by room.
Mean L5 success is 45.49\% with study support (17 tasks), 34.96\% with
living-room support (27 tasks), and 15.43\% with kitchen support
(46 tasks). Several study tasks share the book and caddy, including
back-compartment placement in a different study layout. These results
suggest that support can convey useful scene and manipulation context
even when its demonstrated goal differs from the query. The room
comparison does not isolate background from shared objects and actions.

\section{LIBERO Baselines and Diagnostics}
\label{sec:supp-libero-baselines}

\subsection{WIZARD Reproduction Attempt}
\label{sec:supp-wizard-reproduction}

WIZARD~\cite{bianchi2026wizard} also builds an adapter from one
demonstration of an unseen task. Its generator learns by trying to
predict the adapter weights of experts trained on known tasks. We
implemented this idea from the paper while the authors' code was
awaiting release, using the instruction and video to predict each
source expert's LoRA factors.

One fitted expert succeeded in 46/50 attempts on a source Object
task. On held-out tasks, our generated adapters and the base policy
each scored 0/500 on every suite (0/2,000 per policy). These trials
come from a separate reproduction campaign, with 50 official starts
for each of ten tasks per suite. Decoder layout, token alignment, and
the weight-prediction loss remained open implementation choices. Our
visual tokens also retain language context from joint encoding.
These counts describe the adapters generated by our implementation.

\subsection{DA-MergeLoRA expert selection}
\label{sec:supp-libero-merger-diagnostics}

% Provenance: diagnostic dashboard reviewed on 2026-11-08.
% https://da-mergelora-diagnostics.adhityak.chatgpt.site/
% These results concern the LIBERO implementation, not the published WILDS baseline.
% Add the archived diagnostic configuration/results to the anonymous artifact.

A merger can struggle because its experts are weak or because the
expert it chooses is ill suited to the instruction. We can distinguish
these possibilities on a source task where we already have an expert
trained for that task. We use six validation tasks from the
Spatial-held-out fold. The two Object tasks put chocolate pudding or
orange juice in a basket. The Goal tasks put a bowl on top of a
cabinet or turn on a stove. The Long tasks put cream cheese and butter
in a basket or both moka pots on the stove. Every control uses seed 0, one demonstration,
and the same 50 starts per task, giving 300 trials per condition and
2,400 in total.

\begin{table}[!ht]
  \centering
  \caption{DA-MergeLoRA controls on six LIBERO source-validation tasks.
  Entries are success percentages (successful rollouts/300), from one fold
  and seed. Availability refers to the query task's own source expert.
  The direct and expanded-rank execution paths differ numerically, as
  discussed below.}
  \label{tab:supp-libero-merger-controls}
  \setlength{\tabcolsep}{3.2pt}
  \small
  \begin{tabular}{lcc}
    \toprule
    Control & Own available & Own excluded \\
    \midrule
    Direct own-task expert & 54.33 (163/300) & -- \\
    Forced one-hot own expert & 58.33 (175/300) & -- \\
    Learned merger & 8.00 (24/300) & 1.33 (4/300) \\
    Uniform merger & 0.00 (0/300) & 0.00 (0/300) \\
    Nearest-image expert & 58.33 (175/300) & 0.00 (0/300) \\
    \bottomrule
  \end{tabular}
\end{table}

\paragraph{The stove example.}
The expert trained to turn on the stove succeeds on 49/50 starts.
When it is removed, picking the expert with the most similar images
selects one trained to put a bowl on the stove. It succeeds on 0/50
of those same starts. The scene looks familiar, but the requested
action is different. Across all six tasks, running the matching expert
directly gives 54.33\% success. With that expert removed, the learned
merger gives 1.33\%, and uniform mixing and nearest-image retrieval
each give zero (Table~\ref{tab:supp-libero-merger-controls}).

\paragraph{Reading the controls.}
DA-MergeLoRA fits one expert per source task and trains the merger for
20,000 updates. During training and validation, the sampled task's
own expert is masked. All 30 source experts are available for a held-out
task. On the six validation tasks, giving the merger access to each
task's own expert yields 8.00\% success. Forcing it to use that expert
gives 58.33\%. This last control and direct loading use the
same weights. Their rollout scores differ with rank-480 versus rank-16
BF16 execution. At matched native rank, their outputs agree bit for bit
on the checked inputs. The evidence for difficulty reusing experts
here comes from six tasks in one fold and seed.

\subsection{Controlled blends of two experts}
\label{sec:supp-libero-blending}

What happens when we mix experts trained for different tasks?
Consider two basket tasks. Expert A learned to pick up chocolate
pudding and expert B learned to pick up butter. The robot always
gets the pudding instruction. It succeeds on 25/50 starts with A
alone, 5/50 with a raw half-and-half blend, and 0/50 with B alone.

\begin{figure}[!htb]
  \centering
  \includegraphics[width=\linewidth]{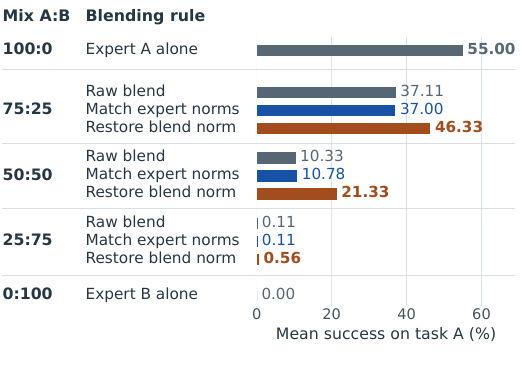}
  \caption{Each bar averages success on A's task over the 18 pairs in
  Table~\ref{tab:supp-libero-blend-pairs} (six tasks, three partners
  each). At 50:50, the pudding--butter pair contributes 5/50 to the
  raw mean of 10.33\%.}
  \label{fig:supp-libero-blends}
\end{figure}

\begin{center}\begin{minipage}{\columnwidth}
  \centering
  \captionof{table}{The 18 expert pairs averaged in
  Fig.~\ref{fig:supp-libero-blends}. All six tasks share the same
  lowest-ranked partner, the expert trained to open the middle drawer.}
  \label{tab:supp-libero-blend-pairs}
  \footnotesize
  \setlength{\tabcolsep}{2.5pt}
  \begin{tabular}{@{}lll@{}}
    \toprule
    \shortstack[l]{Task scored\\(expert A)} &
    \shortstack[l]{Middle partner\\(expert B)} &
    \shortstack[l]{Closest partner\\(expert B)} \\
    \midrule
    \shortstack[l]{Chocolate pudding\\in basket} &
    \shortstack[l]{White mug on plate,\\pudding right of plate} &
    Butter in basket \\[3pt]
    Orange juice in basket & Cream cheese in bowl &
    \shortstack[l]{Salad dressing\\in basket} \\[3pt]
    Bowl atop cabinet & \shortstack[l]{Chocolate pudding\\in basket} &
    Bowl on plate \\[3pt]
    Turn on stove & \shortstack[l]{Wine bottle\\atop cabinet} &
    \shortstack[l]{Turn on stove,\\place moka pot on it} \\[3pt]
    \shortstack[l]{Cream cheese and\\butter in basket} &
    \shortstack[l]{White mug on left plate,\\two-tone mug on right plate} &
    \shortstack[l]{Cream cheese\\in basket} \\[3pt]
    \shortstack[l]{Both moka pots\\on stove} & \shortstack[l]{Wine bottle\\atop cabinet} &
    \shortstack[l]{Turn on stove,\\place moka pot on it} \\
    \bottomrule
  \end{tabular}
\end{minipage}\end{center}

We use six source-validation tasks from the Spatial-held-out fold
(seed~0). For each task, A is its own LoRA expert. Before evaluation,
we rank the other 29 experts by the cosine similarity of their weight
updates to A's across all adapted layers. The lowest, middle, and
highest ranked each serve as B (Table~\ref{tab:supp-libero-blend-pairs}).
B's share before rescaling ranges from 0 to 100\% in steps of 25
percentage points. We use the same 50 starts for each task and
rank~32 throughout. Shared endpoints give 204
distinct conditions and 10,200 trials.

The raw blend mixes the updates directly. Matching expert norms
first scales B's update to the size of A's. Mixing can shrink the
resulting update. \emph{Restore blend norm} first matches B's update
size to A's, mixes the two updates, and then rescales the mixture to
A's original update size. It leads at all three intermediate mixes, reaching
21.33\% at equal shares versus 10.33\% for raw blending. A alone
scores highest overall (55.00\%). Every B alone scores zero on A's task
(Fig.~\ref{fig:supp-libero-blends}).